\pdfoutput=1

\documentclass[11pt]{article}

\usepackage[preprint]{arr_latex/acl}

\usepackage{times}
\usepackage{latexsym}
\usepackage{subcaption}

\usepackage[T1]{fontenc}

\usepackage[utf8]{inputenc}

\usepackage{microtype}

\usepackage{inconsolata}

\usepackage{graphicx}
\usepackage{longtable}
\usepackage{xtab}
\usepackage{array}
\definecolor{forestgreen}{rgb}{0.13, 0.55, 0.13}
\definecolor{brickred}{rgb}{0.8, 0.25, 0.33}

\usepackage{listings}
\usepackage[normalem]{ulem}
\usepackage{microtype}
\usepackage{hyperref}
\usepackage{url}
\usepackage{booktabs}
\usepackage{graphicx}
\usepackage{xspace}
\usepackage{enumitem}
\usepackage{amsmath}
\usepackage{amsfonts}
\usepackage{wrapfig}
\usepackage{multicol}
\usepackage{multirow}
\usepackage{lipsum}
\usepackage{todonotes}
\usepackage{subcaption}
\usepackage{tabularx}
\usepackage{tcolorbox}

\newcommand{\gpt}{\texttt{gpt-4o-2024-08-06}\xspace}
\newcommand{\gptmini}{\texttt{gpt-4o-mini}\xspace}
\newcommand{\llamasmall}{\texttt{LLaMA-3.1-8B-Instruct}\xspace}
\newcommand{\llamalarge}{\texttt{LLaMA-3.1-70B-Instruct}\xspace}
\newcommand{\llamav}{\texttt{LLaMA-3.2-11B-Vision-Instruct}\xspace}
\newcommand{\ministral}{\texttt{Ministral-8B-Instruct}\xspace}
\newcommand{\mistralnemo}{\texttt{Mistral-Nemo-Instruct}\xspace}
\newcommand{\gemmasmall}{\texttt{gemma-2-9B-IT}\xspace}
\newcommand{\gemmalarge}{\texttt{gemma-2-27B-IT}\xspace}
\newcommand{\pixtral}{\texttt{Pixtral-12B-2409}\xspace}

\title{{\texttt{INTERACT}}: Enabling Interactive, Question-Driven Learning in Large Language Models}

\author{Aum Kendapadi\thanks{\hspace{0.5em}Equal contribution} \quad Kerem Zaman\footnotemark[1] \quad Rakesh R. Menon\footnotemark[1] \quad Shashank Srivastava\\
UNC Chapel Hill\\
\texttt{aumken@alumni.unc.edu, \{kzaman, rrmenon, ssrivastava\}@cs.unc.edu}
}

\begin{document}

\maketitle

\begin{abstract}

Large language models (LLMs) excel at answering questions but remain passive learners—absorbing static data without the ability to question and refine knowledge. This paper explores how LLMs can transition to interactive, question-driven learning through student-teacher dialogues. We introduce \texttt{INTERACT} (\texttt{INTER}active learning for \texttt{A}daptive \texttt{C}oncept \texttt{T}ransfer), a framework in which a ``student” LLM engages a ``teacher” LLM through iterative inquiries to acquire knowledge across 1,347 contexts, including song lyrics, news articles, movie plots, academic papers, and images. Our experiments show that across a wide range of scenarios and LLM architectures, interactive learning consistently enhances performance, achieving up to a 25\% improvement, with `cold-start' student models matching static learning baselines in as few as five dialogue turns. Interactive setups can also mitigate the disadvantages of weaker teachers, showcasing the robustness of question-driven learning. \footnote{Our code and dataset are available at \href{https://github.com/aumken/interact}{https://github.com/aumken/interact}.}
\end{abstract}

\section{Introduction}
\label{sec:introduction}
\begin{figure*}[t]
    \centering
    \includegraphics[width=1.0\textwidth]{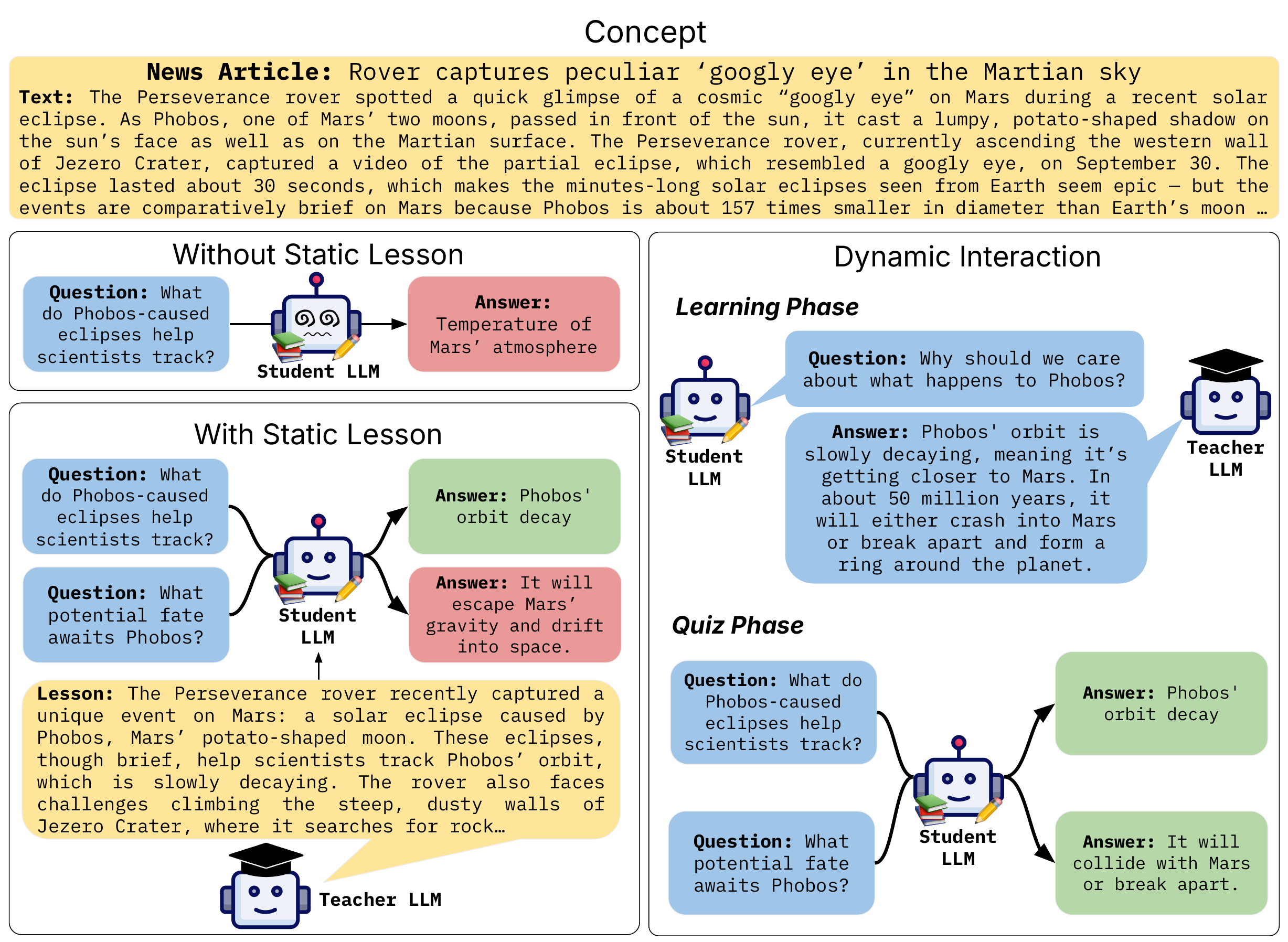}
    \caption{Overview of the \texttt{INTERACT} framework for concept learning in LLMs. Given a new concept, here, a news article from a time period outside of the LLM pretaining data, non-interactive approaches such as zero-shot prompting (\textit{left-top}) and static lessons (\textit{left-bottom}) fail due to lack of information or intricacies in the concept. Through dynamic interaction with a teacher (\textit{right}), a student can learn about a concept more comprehensively.
    }
    \label{fig:intro}
\end{figure*}

Large language models (LLMs) are impressive creatures. They have become fluent at summarizing texts, assisting users, and tackling complex reasoning problems. Yet, their training remains largely static\footnote{Despite alignment methods such as preference tuning \cite{ouyang2022training, rafailov2024direct}.}, relying on fixed datasets rather than interactive processes. In contrast, humans naturally refine their understanding by asking questions, prodding teachers, and poking holes in their explanations until the world makes sense — strategies that help them learn new concepts effectively (see Figure~\ref{fig:intro})~\cite{vygotsky1978mind, ram1991theory}.

Infusing LLMs with this kind of interactive, question-driven inquiry can be valuable for knowledge-intensive domains. Instead of passively absorbing data, an LLM could engage in a dialogue: requesting clarifications, seeking missing details, and testing its evolving comprehension. In education, for example, an AI ``student” could interact with a ``teacher” model, focusing on a learner’s trouble spots rather than delivering the same summary to everyone. In professional contexts such as medicine or scientific research, iterative questioning would let AI systems refine diagnoses, refine hypotheses, and illuminate overlooked in collaboration with human experts.

To advance this vision, we introduce \texttt{INTERACT} (\texttt{INTER}active learning for \texttt{A}daptive \texttt{C}oncept \texttt{T}ransfer), a framework that simulates teacher-student dialogues for LLM-based interactive learning. We ask the question: \emph{How effectively can LLMs learn new concepts through conversational interactions?} Instead of simply consuming summarized information, the ``student” model actively questions the ``teacher”, iteratively building knowledge through inquiry. Much as a human learner hones understanding through persistent, well-placed questions, the student LLM can surface ambiguities, verify assumptions, and guide the conversation toward deeper conceptual clarity.

We evaluate \texttt{INTERACT} on 1,347 unseen contexts spanning movie plots, academic papers, news articles, song lyrics, and visual descriptions—carefully curated to exclude pretraining overlap. This ensures that the student LLM must genuinely acquire new information rather than rely on memorized patterns. We compare two teaching modes: \textit{static lessons}, where a teacher provides a condensed summary of key content, and \textit{dynamic interactions}, where a student LLM engages by asking questions. We evaluate the effectiveness of these approaches by simulating student-teacher interactions.
To measure student learning, we test the LLM's understanding with a quiz either after the static lesson or after each dialogue turn in the dynamic interaction setting. The latter enables a study of LLM questions that most help understanding new concepts. We find that conversational interactions consistently enhance learning achieving \textbf{up to a 25\% improvement in quiz performance}, sometimes matching learning from static data in as few as five dialogue turns. Dynamic interactions also mitigate the advantages of stronger teachers. However, despite the benefits of interaction, student models still underperform teacher LLMs, highlighting the need for improved dialogue strategies.

\texttt{INTERACT}’s implications are wide-ranging. In education, interactive AI tutors could probe students’ knowledge, helping instructors pinpoint misconceptions and personalize their teaching. The same approach can guide trainees in developing their own questioning strategies or assist domain experts in fields like medicine or research by unearthing critical missing links. Rather than functioning as static encyclopedias, LLMs can become proactive collaborators. Our contributions are: 

\begin{itemize}[noitemsep]
    \item A framework for enabling LLMs to learn concepts via teacher-student interactions.
    \item A benchmark of 1,347 unseen contexts, spanning movie plots, academic papers, news articles, song lyrics, and images.
    \item Empirical evidence showing that interactive learning improves concept acquisition.
    \item Analyses showing the importance of adaptive questioning and uncovering insights to improve interactive learning strategies.
\end{itemize}

\section{Related Work}
\label{sec:related_work}
\paragraph{Conversational Machine Learning.}  
Early work on language-guided machine learning often used single-turn instructions and limited examples \citep{srivastava2017joint, hancock-etal-2018-training, arabshahiconversational, menon-etal-2022-clues}. To improve comprehension, researchers have explored active learning \citep{collins2008towards, tamkin2022active} and language-based clarifications \citep{rao-daume-iii-2018-learning, srivastava-etal-2019-learning}. Our approach follows the latter, but unlike methods that rely on templated question generation, we let LLMs produce contextually relevant queries. Our teacher-student paradigm is consonant with knowledge distillation \citep{hinton2015distilling}, though here the ``student” acquires knowledge by active questioning rather than passively receiving information. Our approach is also related to work on document-grounded dialogue~\cite{feng2020doc2dial,feng2021multidoc2dial}, which model multi-turn interactions grounded in text. 

\paragraph{Interactive Learning with LLMs.}  
Recent work shows that LLMs benefit from explanations \citep{wei2022chain, lampinen-etal-2022-language}, self-generated feedback \citep{madaan2024self, chen2024teaching}, and tuning on larger-model outputs \citep{ho-etal-2023-large}. While these focus on teacher-provided information, we highlight the student’s role in shaping the discourse. Other studies examine learning human preferences via dialogue \citep{li2023eliciting, handa2024bayesian} or conversational QA \citep{abbasiantaeb2024let}, but our focus is on student-driven inquiry and eliciting deeper explanations.
\paragraph{Adaptive Learning.}  
Adaptive teaching often involves detecting misconceptions and customizing examples \citep{ross2024adapt}. In contrast, we emphasize a student-led approach, where queries guide the interaction to resolve uncertainties. Our work aligns with benchmarks like MediQ \citep{li2024mediq}, which use multi-turn questioning, but we extend our analyses beyond the medical domain. By allowing student-driven inquiry across varied contexts, we examine how proactive dialogue can enable effective learning.

\paragraph{Interactive Learning in Human Tutoring.}  
Evidence from human learning shows that interactive settings consistently outperform passive instruction. Adaptive tutoring through feedback and dialogue yields substantial gains over classroom-based teaching \citep{bloom1984}. These improvements stem not only from personalization, but from the structure of interaction itself, specifically, student-led questioning and responsive explanation \citep{vanlehn2011}. Our setup parallels this structure: the student model initiates queries to target its own uncertainties. Learning emerges through focused, multi-turn exchanges rather than static instruction.

\section{Experimental Setup}
\label{sec:experiment_setup}
In this section, we delineate our problem setup (\S ~\ref{sec:problem_setup}), outline the creation of our datasets (\S ~\ref{sec:datasets}), present the different interaction scenarios (\S ~\ref{sec:methods}), outline the different models evaluated and our evaluation metric (\S ~\ref{sec:models}). 

\subsection{Problem Setup}
\label{sec:problem_setup}

In this work, a \emph{concept} refers to a distinct unit of knowledge that captures ideas or information embedded in documents across various domains such as literature, sciences, and current world events. Practically, each concept is instantiated through a context document. For example, the concept of a given movie is represented by its Wikipedia plot, while a scientific concept is captured by a research paper excerpt. Our goal is to explore how a student LLM, ($\mathcal{S}$), can learn such concepts by interacting with a teacher LLM ($\mathcal{T}$). 

The student $\mathcal{S}$ can ask any open-ended or information-seeking questions about a concept, while the teacher $\mathcal{T}$ has direct access to the ground-truth context for the concept which it can use to answer those questions. Although one might envision human experts as teachers, large-scale experimentation is more feasible with LLM teachers that can faithfully convey the necessary information. By equipping 
$\mathcal{T}$ with the source context and 
$\mathcal{S}$ with only the answers 
$\mathcal{T}$ provides, we isolate the effects of interactive, inquiry-driven learning.\footnote{Listing \ref{lst:teacher_prompt} shows the instructions provided to the teacher while answering student questions.}

\subsection{Datasets}
\label{sec:datasets}
\begin{table*}[ht]
    \centering
    \scalebox{0.9}{
    \begin{tabular}{l|l|l|l}
        \toprule
        \textbf{Context} & \textbf{Source} & \textbf{\# Contexts} & \textbf{Focus} \\ 
        \midrule
        Song Lyrics & \href{https://genius.com}{Genius} & 467 & 
        Learning from figurative language\\
        News Articles & \href{https://cnn.com}{CNN} & 346 & 
        Learning factual knowledge\\
        Movie Plots & \href{https://www.wikipedia.org}{Wikipedia} & 214 & Learning story elements: characters, events \\ 
        Academic Papers & \href{https://arxiv.org}{arXiv} & 170 & 
        Learning technical knowledge across disciplines \\
        Images & \href{https://cocodataset.org}{COCO} & 150 & Learning to analyze visual contexts \\ 
        \bottomrule
    \end{tabular}
    }
    \caption{Overview of the different content domains used for evaluation, including the number of contexts, sources, and primary focus areas. Each context type tests distinct capabilities of LLMs.
    }
    \label{tab:context_overview}
\end{table*}

Since LLMs gain extensive world knowledge from their pre-training on open web-text \citep{roberts-etal-2020-much}, evaluating their learning abilities on concepts within their pre-training data can lead to ambiguous interpretations. To ensure a robust analysis of concept acquisition, we compiled datasets comprising a range of concepts that are previously unseen by the LLMs. For this, we both automatically scraped and manually compiled song lyrics, movie plots, news articles, academic papers and images, all from after December 2023 (since we tested LLMs pre-trained on data obtained before this period). These documents were collected from platforms such as \href{https://genius.com}{Genius}, \href{https://www.wikipedia.org}{Wikipedia}, \href{https://cnn.com/}{CNN}, \href{https://arxiv.org}{arXiv} and \href{https://cocodataset.org}{COCO} \citep{lin2014microsoft}. This carefully curated dataset spans a range of complexity and information types, enabling a robust evaluation of LLMs' interactive learning performance across various scenarios.

\paragraph{Concepts Dataset Composition} Our evaluation dataset comprises of 1,347 contexts spanning multiple domains. This compilation includes song lyrics, news articles, movie plots, academic papers and images. Table \ref{tab:context_overview} provides an overview of the domains and the number of concepts per domain in the dataset. Further details about the dataset composition can be seen in Appendix~\ref{app:dataset_creation}. 

\paragraph{Static Lesson Generation for Concepts.} We create a `static lesson' for each concept in our dataset by providing \gpt with the context document for the concept, and prompting it to produce a lesson. 
The generated lesson serves as an initial information source the student might leverage during certain interaction scenarios described in \S ~\ref{sec:methods}. Appendix~\ref{app:prompt_templates_static} provides prompts for lesson generation and samples of lessons.

\paragraph{Quiz Generation for Concepts} To measure learning performance, we generate a nine-question quiz per concept, with three levels of difficulty and three questions per level. Questions were crafted using  \texttt{gpt-4o-2024-08-06}. We used an adversarial filtering strategy to exclude questions that could be answered by a \texttt{gpt-4o-mini} model without reference to the provided context. This ensured that each question required eliciting concept information from the provided context. A manual analysis of a random subset of questions by the authors showed that 97\% of them satisfied three criteria: (a) good for testing student understanding, (b) answerable using the context, and (c) do not require deeper knowledge beyond the context. Appendix~\ref{app:dataset_creation} includes details of the quiz generation process. 

\subsection{Student-Teacher Interaction Scenarios}
\label{sec:methods}
In this work, we explore three scenarios to assess the conversational learning capabilities of LLMs: 

\begin{enumerate}[topsep=0pt, leftmargin=*, noitemsep]
    \item \textbf{Static Student with Lesson:} The student only receives the static lesson (no dialogue) before answering quiz questions.
    \item \textbf{Dynamic Student without Lesson:} The student begins with no prior knowledge and acquires information by asking questions.
    \item \textbf{Dynamic Student with Lesson:} The student first receives the static lesson, then refines understanding through questions.
\end{enumerate}

These scenarios let us investigate whether interactive questioning can complement or surpass static instruction, and how the quality of teacher and lesson information shapes learning outcomes.
We explore five research questions:

\begin{itemize} 
    \item \textbf{RQ1:} How well can students learn concepts from static lessons?
    \item \textbf{RQ2:} How well can students learn concepts through interactions?
    \item \textbf{RQ3:} How does the quality of the teacher and lesson affect dynamic student performance?
    \item \textbf{RQ4:} Can borrowed interactions improve student performance?
    \item \textbf{RQ5:} What patterns or features emerge in the questions generated by the student model?
\end{itemize}

\subsection{Models and Evaluation Metrics}
\label{sec:models}
\paragraph{Models.} 
We evaluate a range of open and closed-source LLMs as both teachers and students. For text-based domains, we test \gptmini and instruction tuned versions of \texttt{LLaMA-3.1-8B/70B}, \texttt{Ministral-8B}, \texttt{Mistral-Nemo}, and \texttt{gemma-2-9B/27B}. For the image domain, we experiment with \gptmini, \pixtral, and \llamav (multi-modal LLMs). These models are chosen for their strong language understanding and generation capabilities, which are vital for conversational learning. Unless mentioned otherwise, we provide the student model with the \texttt{gpt-4o} generated lesson in the static and dynamic settings.\footnote{We omit evaluations with \texttt{gpt-4o} in the student-teacher setup primarily due to the high cost, which made extensive experimentation in dynamic settings infeasible.} 

\begin{figure}[h!]
    \centering
    \includegraphics[width=1.0\linewidth]{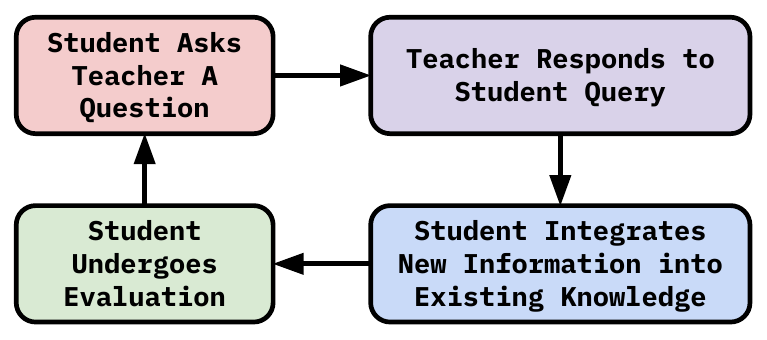}
    \caption{Workflow of the evaluation process in each dialogue turn of the \textit{dynamic interactions} setting, starting with the student asking a concept-related question. 
    }
    \label{fig:workflow}
\end{figure}
During dynamic interactions, after every dialogue turn, the student model is prompted to integrate the newly acquired information from the ongoing conversation and, when applicable, the prior static lesson. This integration is done through appending the conversation history to the student's context, allowing the model to consolidate knowledge incrementally. 
The student then uses this updated context to answer quiz questions, as illustrated in Figure~\ref{fig:workflow}. For fair comparison, dynamic dialogues are generated with a temperature value set to $1.0$, while quiz answers are generated with a temperature of $0$. All experiments are repeated across three seeds to ensure robustness.

Our metric for measuring concept learning performance is the accuracy of the student model's responses in concept quizzes, measured as the fraction of quiz questions answered correctly. This metric quantifies how well the student has internalized the concept discussed during the interactions. 

\section{Results}
\label{sec:experiments}
In this section, we present our findings. We begin by establishing a baseline for non-interactive scenarios (\S\ref{sec:non_interactive_results}) before exploring how interactive questioning affects concept learning (\S\ref{sec:interactive_results}). Subsequently, we analyze the role of teacher and lesson quality (\S\ref{sec:teacher_lesson_quality}), investigate whether borrowed interactions can substitute for active engagement (\S\ref{sec:borrowed_interactions}), and consider what factors correlate with successful conversational learning (\S\ref{sec:factors}).

\subsection{RQ1: How Well Can Students Learn Concepts from Static Lessons?}
\label{sec:non_interactive_results}
\begin{figure}[t]
    \centering
    \includegraphics[width=1.0\linewidth]{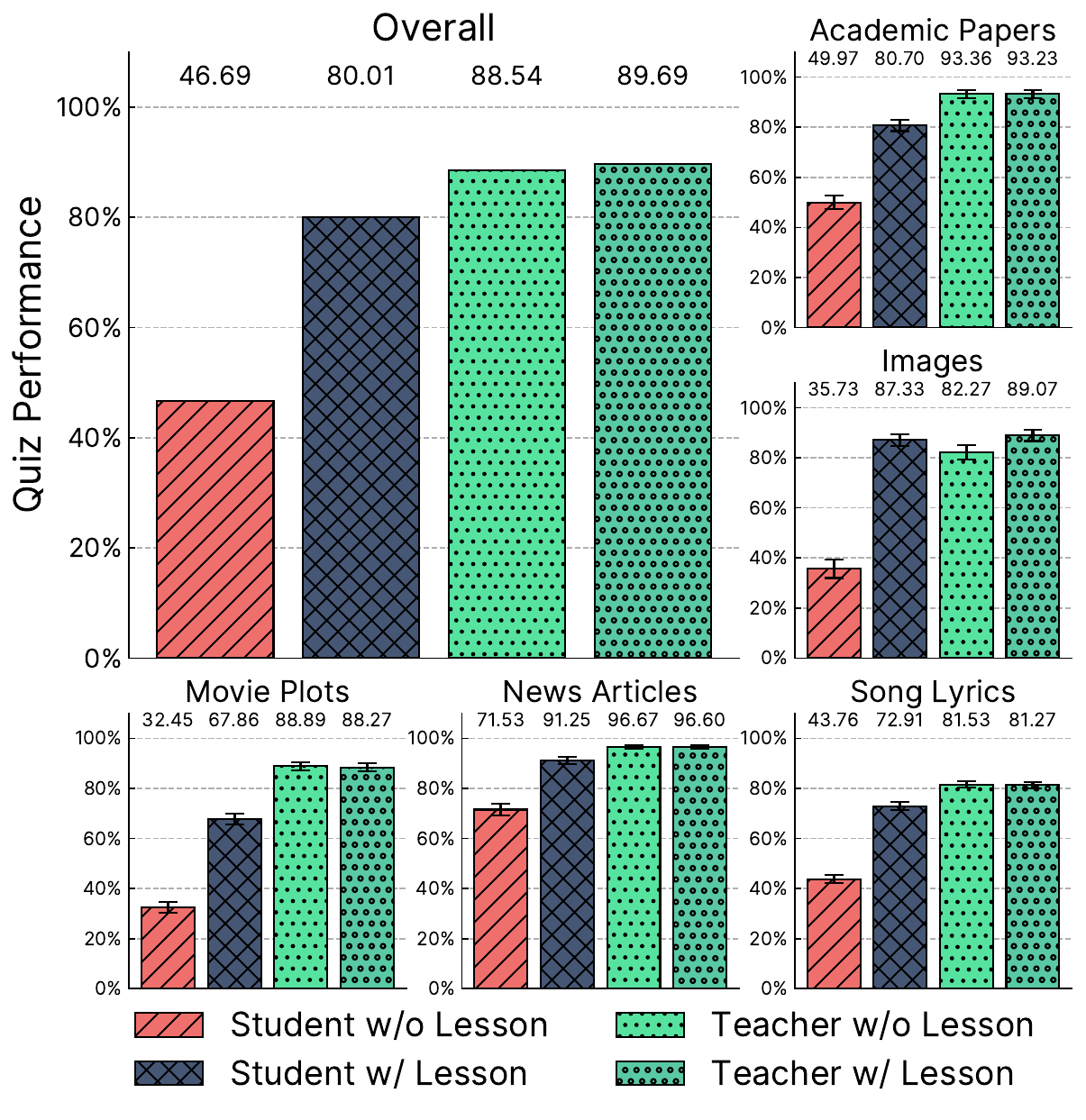}
    \vspace{-1em}
    \caption{Average quiz performance of student and teacher \gptmini models across different domains. Errorbars indicate the 95\% confidence interval calculated by bootstrap. Quiz performance with other LLMs can be found in Appendix Figure \ref{fig:other_llms_static}.}
    \label{fig:static-eval-main}
\end{figure}

We first evaluate students’ ability to learn concepts without interaction. Here, the student either receives (1) no knowledge about the concept, or (2) the \textit{static lesson} for the concept. We then compare these students’ quiz performance to the teacher’s performance. The teacher is given direct access to the context for the concept (and optionally, also the static lesson) and serves as a conceptual upper bound for the student models' performance.

\begin{figure*}
    \centering
    \includegraphics[width=\linewidth]{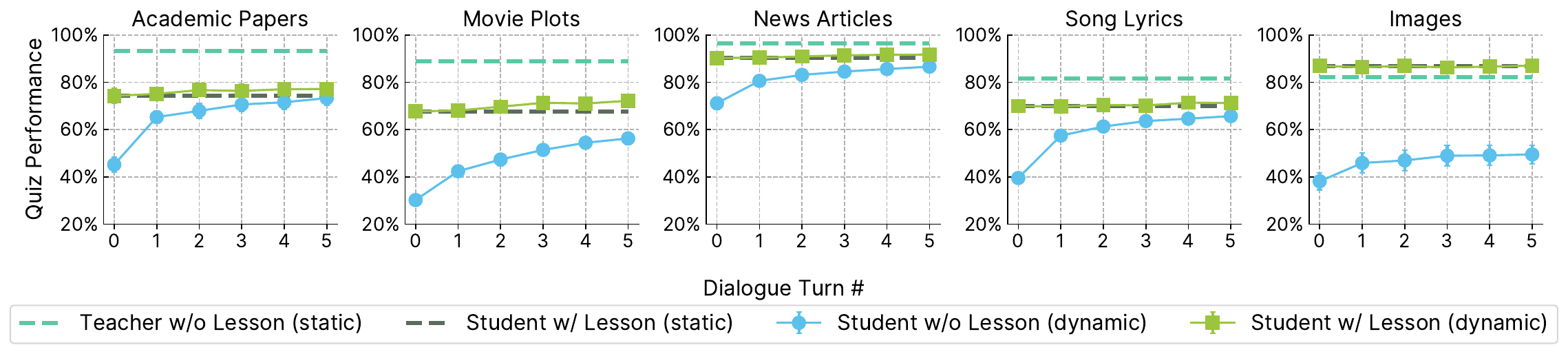}
    \vspace{-1em}
    \caption{Performance of student \gptmini models across various static and dynamic evaluation settings over five interaction rounds. Errorbars indicate the 95\% confidence interval calculated by bootstrap. Performance with other LLMs can be found in App. Fig. \ref{fig:other_llms_dynamic}. 
    }
    \label{fig:dynamic-eval}
\end{figure*}

\begin{table*}[h!]
    \centering
    \scalebox{0.85}{
    \begin{tabular}{lcccccccc}
    \toprule
    Model        & \multicolumn{2}{c}{Student w/o Lesson} & \multicolumn{2}{c}{Student w/ Lesson} & Teacher & \multicolumn{2}{c}{\begin{tabular}{c}Quiz Recovery of\\ Student w/o Lesson (\%)\end{tabular}} \\
    \cmidrule(lr){2-3} \cmidrule(lr){4-5} \cmidrule(lr){7-8}
                 & Start & End (\(\Delta\))       & Start & End (\(\Delta\))       & Performance    & wrt S w/ L Start & wrt Teacher  \\
    \midrule
    gpt-4o-mini  & 47.91 & 73.68 \textcolor{forestgreen}{(+25.77)} & 78.83 & 81.23 \textcolor{forestgreen}{(+2.40)} & 90.05   & 91.23            & 81.47        \\
    \midrule
    LLaMA-8B     & 38.12 & 60.13 \textcolor{forestgreen}{(+22.01)} & 70.88 & 72.43 \textcolor{forestgreen}{(+1.55)} & 85.70   & 84.89            & 75.49        \\
    LLaMA-70B    & 60.34 & 76.81 \textcolor{forestgreen}{(+16.47)} & 80.58 & 82.94 \textcolor{forestgreen}{(+2.36)} & 91.24   & 93.81            & 85.13        \\
    \midrule
    Ministral-8B & 33.82 & 59.66 \textcolor{forestgreen}{(+25.84)} & 67.68 & 69.36 \textcolor{forestgreen}{(+1.68)} & 82.38   & 85.14            & 72.33        \\
    Mistral-Nemo & 46.81 & 70.61 \textcolor{forestgreen}{(+23.80)} & 74.06 & 76.82 \textcolor{forestgreen}{(+2.76)} & 82.05   & 91.81            & 81.90        \\
    \midrule
    Gemma-9B     & 47.19 & 63.74 \textcolor{forestgreen}{(+16.55)} & 75.39 & 77.08 \textcolor{forestgreen}{(+1.69)} & 89.15   & 83.21            & 71.75        \\
    Gemma-27B    & 52.83 & 65.17 \textcolor{forestgreen}{(+12.34)} & 77.77 & 79.58 \textcolor{forestgreen}{(+1.81)} & 89.83   & 86.72            & 73.47        \\
    \bottomrule
    \end{tabular}
    }
    \caption{Aggregated performance for LLMs across text domains, showing average start/end performances and recovery percentages for students without/with lessons in the dynamic setting and the static teacher performance. Higher recovery percentages indicate stronger concept understanding as measured by quiz performance.}
    \label{tab:dynamic_main_results}
\end{table*}

\paragraph{Results.} 
Figure \ref{fig:static-eval-main} shows the quiz performance of student and teacher \gptmini models across various domains from our dataset, based on their access to information (we see similar trends across other LLMs; see Appendix \ref{app:additional_results}). Students provided with no knowledge of new concepts perform above chance, likely relying on pre-training knowledge on the Academic Papers and News Articles domains. 
This is likely because new concepts build upon previously established theories and facts, respectively, for these domains. When provided with a \textit{static lesson}, student performance improves significantly ($p < 0.01$\footnote{We use the student t-test to perform the hypothesis testing across all experiments.}) across all domains, though it remains significantly lower ($p < 0.01$) than that of teachers with concept context access in all domains except Images. For this domain, the student with static lesson performs better than the teacher without the lesson, and we posit this is because of the stronger text-based training of \gptmini compared with its image-processing abilities. This trend does not hold with other vision language models like \pixtral (Figure \ref{fig:pixtral-static}). As would be expected, for the text-based domains, teacher performance with or without the static lesson remains largely the same (<1\% difference on average across text domains).

\subsection{RQ2: How Well Can Students Learn Concepts through Interactions?}
\label{sec:interactive_results}
Next, we analyze the accuracy of concept learning when a student interactively engages in dialogue with a teacher. We track the learning progress over five rounds of interactions. We compare this to two approaches: (1) a student model that receives a static lesson from the teacher without interaction, and (2) the teacher model that has full knowledge of the concept, which serves as a conceptual upper bound. Notably, exceptions to this upper bound in the Images domain occur for \llamav and \gptmini, where the student with a static lesson outperforms the teacher.

\paragraph{Main Results.} Figure \ref{fig:dynamic-eval} shows that students without a lesson initially lag behind their static-lesson counterparts, but improve substantially after each question-answer exchange for \gptmini. These improvements hold across different language models (+22\% absolute gain for LLaMA-8B, Table \ref{tab:dynamic_main_results}; Figure \ref{fig:other_llms_dynamic} in Appendix \ref{app:additional_results}). Despite steady improvement, most dynamic students do not surpass the static-lesson baseline within five rounds, particularly in text domains. Longer or more strategically guided interactions might close the gap further. In the Images domain, however, performance quickly saturates over the first couple of interaction rounds (see Figure \ref{fig:other_vllms_dynamic} for \texttt{Pixtral-12B} and \texttt{LLaMA-3.2-11B} results).

These results suggest that LLMs can generate comprehensive and domain-relevant questions during the conversation to learn about new concepts, with the potential for further refinement with more interaction rounds (see Figure \ref{fig:other_vllms_dynamic_longer}). Table \ref{tab:sampled_questions_gpt4o_mini}  shows examples of questions generated by the student \gptmini during conversations in the student without a static lesson setup. Additional examples for other student models are provided in Table \ref{tab:sampled_questions}. 

\begin{table}
    \centering
    \small
    \begin{tabular}{@{}llp{3.5cm}}
    \toprule
    \textbf{Domain} & \textbf{Round} & \textbf{Question} \\
    \midrule
    Song Lyrics & round 1 & What inspired Kylie Minogue to create "Hold On To Now," and  what themes does the song explore? \\
    Academic Papers & round 2 & How do Convolutional Neural Networks (CNNs) process seismic  data differently than traditional data analysis methods? \\
    News Articles & round 3 & What are the key legal arguments the district attorney has  used in his previous clemency requests in other cases, if  any? \\
    Movie Plots & round 5 & What are some specific  scenes or moments from "Problemista"  that exemplify Alejandro's unwavering determination in the  face of adversity? \\
    \bottomrule
    \end{tabular}
    \caption{Samples of Student Questions from Interactions with Teachers for \gptmini.} \label{tab:sampled_questions_gpt4o_mini}
\end{table}

When starting from a static lesson, adding interaction to the student leads to statistically significant improvements ($p < 0.01$) over the student without a lesson. However, student performance remains significantly lower ($p < 0.01$) than that of teachers in all domains except Images, indicating scope for better interaction strategies. In summary, \emph{while students are capable of learning through interaction, the extent of knowledge acquired via this method significantly lags behind that of teachers}.

\subsection{RQ3: Does Teacher and Lesson Quality Influence Dynamic Learning?}
\label{sec:teacher_lesson_quality}
We now examine whether improving teacher quality or the initial lesson can enhance interactive learning. Are students paired with stronger teachers or given higher-quality initial lessons better poised to reach teacher-level understanding?

\paragraph{Study Design} 
To evaluate the effects of both lesson quality and teacher strength, we design two complementary experiments:
\begin{enumerate}[itemsep=0em, leftmargin=1.2em]
\item	\textbf{Effect of Static Lesson Quality:} We substitute the static \gpt-generated lesson provided to a \texttt{LLaMA-8B} student with a lesson generated by a \texttt{LLaMA-8B} teacher. We then measure the student model’s concept learning accuracy under two conditions: (a) static (no interaction) and (b) dynamic (after five interaction rounds with the \texttt{LLaMA-8B} teacher, building upon the initial static lesson).
\item	\textbf{Effect of Teacher Strength:} We pair a \texttt{LLaMA-8B} student with a stronger \texttt{LLaMA-70B} teacher, which also provides the static lesson. To analyze learning behavior with a weaker teacher, we reverse the setup and have \texttt{LLaMA-70B} acts as the student and \texttt{LLaMA-8B} as the teacher.
\end{enumerate}

\begin{table}[]
    \centering
    \scalebox{0.75}{
    \begin{tabular}{l|c|c|c|c}
    \toprule
    \textbf{Context} & \multicolumn{2}{c}{\textbf{gpt-4o Lesson}} & \multicolumn{2}{c}{\textbf{LLaMA-8B Lesson}} \\
    \cmidrule(lr){2-3} \cmidrule(lr){4-5}
     & \textbf{Static} & \textbf{Post-Int.} & \textbf{Static} & \textbf{Post-Int.} \\
    \midrule
    Academic Papers & 73.02 & 71.80 & 70.18 & 72.59 \\
    Movie Plots & 58.68 & 61.69 & 51.03 & 60.97 \\
    News Articles & 87.18 & 88.20 & 81.29 & 88.43 \\
    Song Lyrics & 64.65 & 64.63 & 59.20 & 64.65 \\
    \bottomrule
    \end{tabular}
    }
    \caption{Quiz performance comparison for \llamasmall in the static and dynamic (Post-Interaction) settings using gpt-4o and \llamasmall static lessons.}
    \label{tab:static_lesson_ablation}
\end{table}

\paragraph{Main Results.}
From Table~\ref{tab:static_lesson_ablation}, we find that a stronger teacher’s static lesson (\gpt) confers a 3-5\% average improvement in static student performance compared to the \llamasmall-generated lesson. However, after five interaction rounds, the difference narrows considerably, with final scores differing by about 1\%. This suggests that while a high-quality initial lesson can boost static accuracy, the dynamic interaction process largely mitigates any initial lesson quality gap.

\begin{table}[h!]
\centering
\scalebox{0.75}{
\begin{tabular}{l|c|c|c|c}
\toprule
\textbf{Student} $\rightarrow$ & \multicolumn{2}{c|}{\textbf{LLaMA-8B}} & \multicolumn{2}{c}{\textbf{LLaMA-70B}} \\ 
\cmidrule{2-5}
\textbf{Teacher} $\rightarrow$ & \textbf{L-8B} & \textbf{L-70B} & \textbf{L-70B} & \textbf{L-8B} \\
\midrule
Aca. Pap. & 71.80 & 71.63 \textcolor{brickred}{(-0.17)} & 83.97 & 84.31 \textcolor{forestgreen}{(+0.34)} \\
Mov. Plts & 61.69 & 61.26 \textcolor{brickred}{(-0.43)} & 76.31 & 76.99 \textcolor{forestgreen}{(+0.68)} \\
News Art. & 88.20 & 87.71 \textcolor{brickred}{(-0.49)} & 93.55 & 93.02 \textcolor{brickred}{(-0.53)} \\
Song Lyr. & 64.63 & 66.16 \textcolor{forestgreen}{(+1.53)} & 77.93 & 76.54 \textcolor{brickred}{(-1.39)} \\
\bottomrule
\end{tabular}
}
\caption{Post-interaction concept quiz performance comparison when using \llamasmall and \llamalarge as teacher models. Note, these are conducted in the student with lesson dynamic setting. (L=LLaMA)}
\label{tab:teacher_ablation}
\end{table}

Surprisingly, there is also a  minimal difference in performance from having a stronger teacher (Table~\ref{tab:teacher_ablation}). A weaker or stronger teacher does not consistently improve the student’s final performance after dynamic interactions. This indicates that simply increasing teacher strength does not guarantee deeper student understanding or more effective question-asking behavior. Students fail to consistently capitalize on a teacher’s superior knowledge through improved questioning strategies.

These suggest that \textit{while stronger lessons and teachers provide a head start, the depth of student-driven inquiry during interactive learning is a key determinant of concept mastery}.

\subsection{RQ4: Can Borrowed Interactions Substitute for Proactive Engagement?}
\label{sec:borrowed_interactions}
Next we explore the following question: \textit{Can weaker students benefit from previously generated, high-quality interaction transcripts from stronger students — effectively ``borrowing” another student’s dialogue—without actively participating?}

\paragraph{Study Design.} We generate transcripts from an interaction between a strong teacher-student pair (both \texttt{LLaMA-70B} models) and provide them as context (similar to the teacher lesson)
to a weaker student (\texttt{LLaMA-8B}) that never engaged in that particular dialog. We then measure if the weaker student’s performance improves from this passive exposure alone.

\begin{table}[h!]
\centering
\scalebox{0.7}{
\begin{tabular}{l|c|c|c|c}
\toprule
\textbf{Eval. LLM} $\rightarrow$ & \multicolumn{2}{c|}{\textbf{LLaMA-8B}} & \multicolumn{2}{c}{\textbf{LLaMA-70B}} \\ 
\cmidrule{2-5}
\textbf{Interactions} $\rightarrow$ & \textbf{L-8B} & \textbf{L-70B} & \textbf{L-70B} & \textbf{L-8B} \\
\midrule
Aca. Pap. & 67.94 & 65.09 (\textcolor{brickred}{-2.85}) & 80.43 & 81.55 (\textcolor{forestgreen}{+1.12}) \\
Mov. Plts & 41.38 & 44.30 (\textcolor{forestgreen}{+2.92}) & 65.98 & 61.83 (\textcolor{brickred}{-4.15}) \\
News Art. & 76.29 & 77.04 (\textcolor{forestgreen}{+0.75}) & 87.57 & 87.47 (\textcolor{brickred}{-0.10}) \\
Song Lyr. & 56.91 & 59.75 (\textcolor{forestgreen}{+2.84}) & 73.26 & 70.66 (\textcolor{brickred}{-2.60}) \\
\bottomrule
\end{tabular}
}
\caption{Concept quiz performance when using \textit{student w/o lesson} interactions of \texttt{LLaMA-8B} and \texttt{LLaMA-70B} as context for \texttt{LLaMA-8B} and \texttt{LLaMA-70B} student models.}
\label{tab:abduction_student_wo_lesson}
\end{table}

\begin{table}[h!]
\centering
\scalebox{0.7}{
\begin{tabular}{l|c|c|c|c}
\toprule
\textbf{Eval. LLM} $\rightarrow$ & \multicolumn{2}{c|}{\textbf{LLaMA-8B}} & \multicolumn{2}{c}{\textbf{LLaMA-70B}} \\ 
\cmidrule{2-5}
\textbf{Interactions} $\rightarrow$ & \textbf{L-8B} & \textbf{L-70B} & \textbf{L-70B} & \textbf{L-8B} \\
\midrule
Aca. Pap.  & 71.80 & 72.42 (\textcolor{forestgreen}{+0.62}) & 83.97 & 84.44 (\textcolor{forestgreen}{+0.47}) \\
Mov. Plts & 61.69 & 61.67 (\textcolor{brickred}{-0.02}) & 76.31 & 76.62 (\textcolor{forestgreen}{+0.31}) \\
News Art. & 88.20 & 87.77 (\textcolor{brickred}{-0.43}) & 93.55 & 93.38 (\textcolor{brickred}{-0.17}) \\
Song Lyr. & 64.63 & 66.43 (\textcolor{forestgreen}{+1.80}) & 77.93 & 77.15 (\textcolor{brickred}{-0.78}) \\
\bottomrule
\end{tabular}
}
\caption{Concept quiz performance when using \textit{student w/ lesson} interactions of \texttt{LLaMA-8B} and \texttt{LLaMA-70B} as context for \texttt{LLaMA-8B} and \texttt{LLaMA-70B} student models.}
\label{tab:abduction_results}
\end{table}

\paragraph{Main Results.} Tables \ref{tab:abduction_student_wo_lesson} and \ref{tab:abduction_results} show that passive exposure to borrowed transcripts does not significantly improve performance.
Conversely, we also observe that observing the interactions of a weaker student-teacher combination does not diminish performance considerably across most scenarios (with the exception of the Movie Plots and Song Lyrics domains in the student without lesson dynamic setting).
This suggests that the benefits of interactive learning are not about conversation quality, but the student’s capacity to appropriately ask questions. Thus, \textit{passive exposure to high-quality content cannot substitute for pro-active engagement}. 

\subsection{RQ5: What Interaction Factors Predict Student Learning Gains?}
\label{sec:factors}
Finally, we examine which aspects of teacher-student interactions drive better learning outcomes. 

\paragraph{Study Design.}
We first collect a comprehensive set of 53 interaction-related features\footnote{We provide the full list of features in Appendix Table \ref{tab:features_regression}.} that can be computed from the teacher-student transcripts. These factors include syntactic features of the student/teacher responses, statistics about tokens, metrics of linguistic complexity, semantic relatedness of questions and responses, among others.  
For each domain and configuration with \texttt{gpt-4o-mini} as student and teacher models, we extract these features from the recorded interactions and aggregate them into a feature matrix. 
We then train a random forest regressor, using best-found hyperparameters from cross-validation, to predict the learning gain of the student model after each round of interaction with a teacher. 
Performance is evaluated using the $R^2$ score on a held-out test set.

\paragraph{Main Results.}
In most domains, the predictive power of our feature set is low: $R^2$ scores on held-out data are often close to zero. However, for the song lyrics domain, $R^2$ scores reach up to 0.14, suggesting that meaningful signals can be extracted in specific conditions. 
The top contributing features are cumulative exposure (number of unique tokens), overlap between quiz questions and student questions, semantic alignment between student and quiz questions, response information density, and response correctness. This suggests that \textit{our current feature sets and metrics are insufficient for robustly capturing the nuanced factors driving interactive learning success}. Identifying richer metrics remains a key challenge.

\section{A Future of Conversational Learning}
\label{sec:conclusion}
Our findings suggests promise for a new paradigm for learning in LLMs: by shifting from static data absorption to interactive, curiosity-driven dialogue. The \texttt{INTERACT} framework and curated dataset provide a fertile testing ground for refining this paradigm, where learning experiences are not delivered by a teacher but co-constructed by learners. Despite the benefits of interactive learning, current student models lag behind teachers, indicating the need for better interaction strategies.

Future work can explore extending existing machine learning theories, such as active learning, to analyze and optimize interactive learning methods. By treating active learning as a special case, these extensions could lead to new theoretical frameworks that capture the complexities of real-time, adaptive learning. They also point to concrete applications: dynamic AI tutors humans with evolving lesson plans, accelerated scientific discovery, and enhanced reasoning with images, audio, or video data. Investigating methods for long-term retention of knowledge acquired through interaction will also be critical. At its heart, this line of research nudges us toward a future where machine learning systems not only learn from us, but learn with us.

\section*{Limitations}
\label{sec:limitations}
Our investigation here has some important limitations. 
A significant limitation lies in our evaluation method. While quiz performance is informative, it may not capture all aspects of concept understanding. This metric might overlook nuanced comprehension or the ability to apply learned concepts in novel contexts. Moreover, our focus on immediate concept acquisition leaves open questions about long-term retention and integration of knowledge gained through interactive learning. More comprehensive evaluation methods could offer a more holistic picture of LLM learning, including assessments of reasoning ability, knowledge transfer, and conceptual integration over time.

The scalability of our approach to larger datasets, longer conversations, or more complex concepts remains untested. As the complexity of tasks increases, the computational resources required for extended dialogues could become prohibitive, potentially limiting practical applicability in real-world settings. This scalability challenge is closely tied to ethical considerations, particularly regarding the deployment of AI in educational contexts. Important issues such as AI transparency, potential biases in learning outcomes, and the impact on human learning processes when interacting with AI teachers remain unaddressed.

\section{Acknowledgements}
This work was supported by the NSF grant DRL2112635. The authors would also
like to thank anonymous reviewers for valuable feedback to improve the manuscript.

\bibliography{custom}

\section*{Appendix}
\appendix
\section{Extended Related Work}
\label{app:extended_related_work}
\paragraph{Knowledge Distillation and Interactive Reasoning.}
Knowledge distillation methods traditionally train student models on outputs generated by teachers, often leading to discrepancies between training and inference data (Kim \& Rush, 2016; Sanh et al., 2019). Generalized Knowledge Distillation (Agarwal et al., 2023) addresses this by incorporating self-generated sequences and teacher feedback. While these approaches align student outputs with teacher feedback, our work evaluates learning through dynamic interactions, focusing on how students refine their knowledge by engaging in conversation and inquiry. Interactive frameworks also highlight the importance of dialogue in improving LLM performance. Studies show that LLMs benefit from human explanations (Wei et al., 2022; Lampinen et al., 2022), self-generated feedback (Madaan et al., 2023; Chen et al., 2024), and fine-tuning on explanations from larger models (Ho et al., 2023). Our approach extends these findings by focusing on the student’s ability to ask informative questions, enabling richer teacher explanations and deeper learning.

\paragraph{Dynamic Question Generation and Simulated Interaction.}
Dynamic question generation methods have demonstrated that pre-trained models can tailor questions based on a student’s knowledge state (Srivastava \& Goodman, 2021). Their LM-KT model personalizes questions to match student proficiency, enhancing learning outcomes compared to static question pools. While this approach centers on teacher-driven question generation, our work focuses on student-driven questioning, where students actively inquire to address their knowledge gaps. Simulated environments further highlight the role of interactive dialogue in assessing LLM capabilities. Frameworks like SOTOPIA (Zhou et al., 2024) and COBLOCK (Wu et al., 2024) evaluate social intelligence and collaboration through multi-turn interactions. These studies emphasize adaptability and communication in achieving shared goals. In contrast, our work explores how student-teacher dialogues facilitate concept learning, emphasizing the student’s role in refining knowledge through inquiry across diverse domains.

\section{Additional Results}
\label{app:additional_results}
\subsection{Interaction Utilization}

As shown in Section~\ref{sec:borrowed_interactions}, passive exposure to high-quality content is insufficient to replace proactive engagement. Specifically, providing transcripts of interactions between a stronger teacher-student pair to a weaker student does not lead to comparable performance. We also observe that interactions involving weaker students do not substantially degrade overall performance. To assess whether models genuinely leverage the observed interactions, we conduct an ablation study in which we replace meaningful interactions with irrelevant ones.

\begin{table}[!ht]
    \centering
    \scalebox{0.7}{
    \begin{tabular}{l|c|c|c|c}
    \toprule
    \textbf{Student} $\rightarrow$ & \multicolumn{2}{c}{\textbf{w/o Lesson}} & \multicolumn{2}{c}{\textbf{w/ Lesson}} \\
    \cmidrule(lr){2-3} \cmidrule(lr){4-5}
    \textbf{Interactions} $\rightarrow$ & \textbf{Orig.} & \textbf{Random} & \textbf{Orig.} & \textbf{Random} \\
    \midrule
        Academic Papers & 78.54 & 47.33 \textcolor{red}{(-31.21)} & 82.56 & 47.17 \textcolor{red}{(-35.39)} \\
    Movie Plots & 57.85 & 34.11 \textcolor{red}{(-23.74)} & 73.94 & 34.21 \textcolor{red}{(-39.73)} \\
    News Articles & 87.40 & 73.11 \textcolor{red}{(-14.29)} & 92.52 & 72.95 \textcolor{red}{(-19.57)} \\
    Song Lyrics & 68.92 & 45.84 \textcolor{red}{(-23.07)} & 74.94 & 45.53 \textcolor{red}{(-29.41)} \\
    Images & 49.47 & 23.73 \textcolor{red}{(-25.73)} & 87.07 & 23.60 \textcolor{red}{(-63.47)} \\
    \bottomrule
    \end{tabular}
    }
    \caption{Quiz performance comparison for \gptmini with its original and random interactions for students without/with lessons. Random interaction results are averaged across 3 seeds.}
    \label{tab:concept_transfer}
\end{table}

\paragraph{Study Design.} For each domain, we sample 100 concept interactions and replace them with interactions from a randomly selected, unrelated concept within the same domain. In this setup, the interactions are sourced from dialogues between \gptmini student and teacher models.

\paragraph{Main Results.} Table~\ref{tab:concept_transfer} shows that exposure to irrelevant transcripts results in a significant performance decline. This indicates that students are indeed making effective use of the observed interactions.

\begin{table*}[htbp]
\centering
\small
\caption{Default hyperparameters for the dynamic setting experiments. Static configuration uses the same hyperparameters with the student quiz evaluation module alone.}
\label{tab:hyperparameters}
\begin{tabular}{@{}p{4cm} p{6cm} p{3cm}@{}}
\toprule
\textbf{Feature Name} & \textbf{Description} & \textbf{Value} \\
\midrule
\multicolumn{3}{l}{\textbf{Experiment Parameters}} \\
\midrule
\textbf{num\_interaction\_rounds} & Total number of interaction rounds. & 5 \\
\textbf{seed} & Random seed for reproducibility. & 0 \\
\midrule
\multicolumn{3}{l}{\textbf{Student Question Generation}} \\
\midrule
\textbf{max\_tokens} & Maximum tokens for question generation. & 256 \\
\textbf{temperature} & Sampling temperature for question generation. & 1 \\
\midrule
\multicolumn{3}{l}{\textbf{Student Quiz Evaluation}} \\
\midrule
\textbf{max\_tokens} & Maximum tokens for quiz evaluation. & 10 \\
\textbf{temperature} & Sampling temperature for quiz evaluation. & 0 \\
\midrule
\multicolumn{3}{l}{\textbf{Student Summary Generation}} \\
\midrule
\textbf{max\_tokens} & Maximum tokens for summary generation. & 256 \\
\textbf{mode} & Mode for summary aggregation. & concat \\
\textbf{temperature} & Sampling temperature for summary generation. & 0.7 \\
\midrule
\multicolumn{3}{l}{\textbf{Teacher Answer Generation}} \\
\midrule
\textbf{lesson\_mode} & Lesson provider. & \gpt \\
\textbf{max\_tokens} & Maximum tokens for answer generation. & 512 \\
\textbf{temperature} & Sampling temperature for answer generation. & 0.7 \\
\bottomrule
\end{tabular}
\end{table*}

\subsection{Additional Static Results}

\begin{figure*}
    \centering
    \begin{subfigure}[b]{0.45\textwidth}
        \centering
        \includegraphics[width=0.9\textwidth]{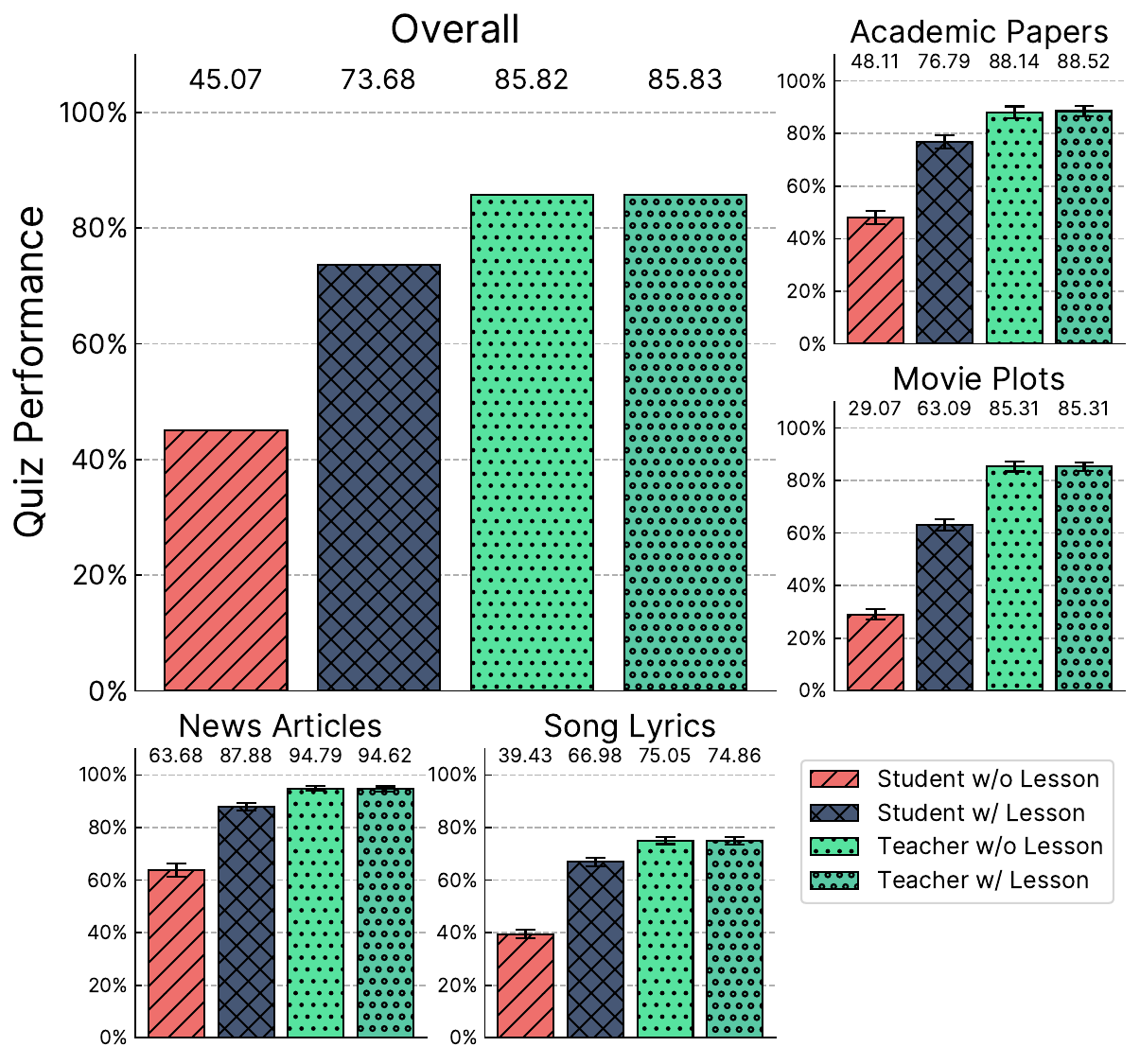} 
        \caption{\bf \large \llamasmall}
        \label{fig:llama-8b-static}
    \end{subfigure}
    \hfill
    \begin{subfigure}[b]{0.45\textwidth}
        \centering
        \includegraphics[width=0.9\textwidth]{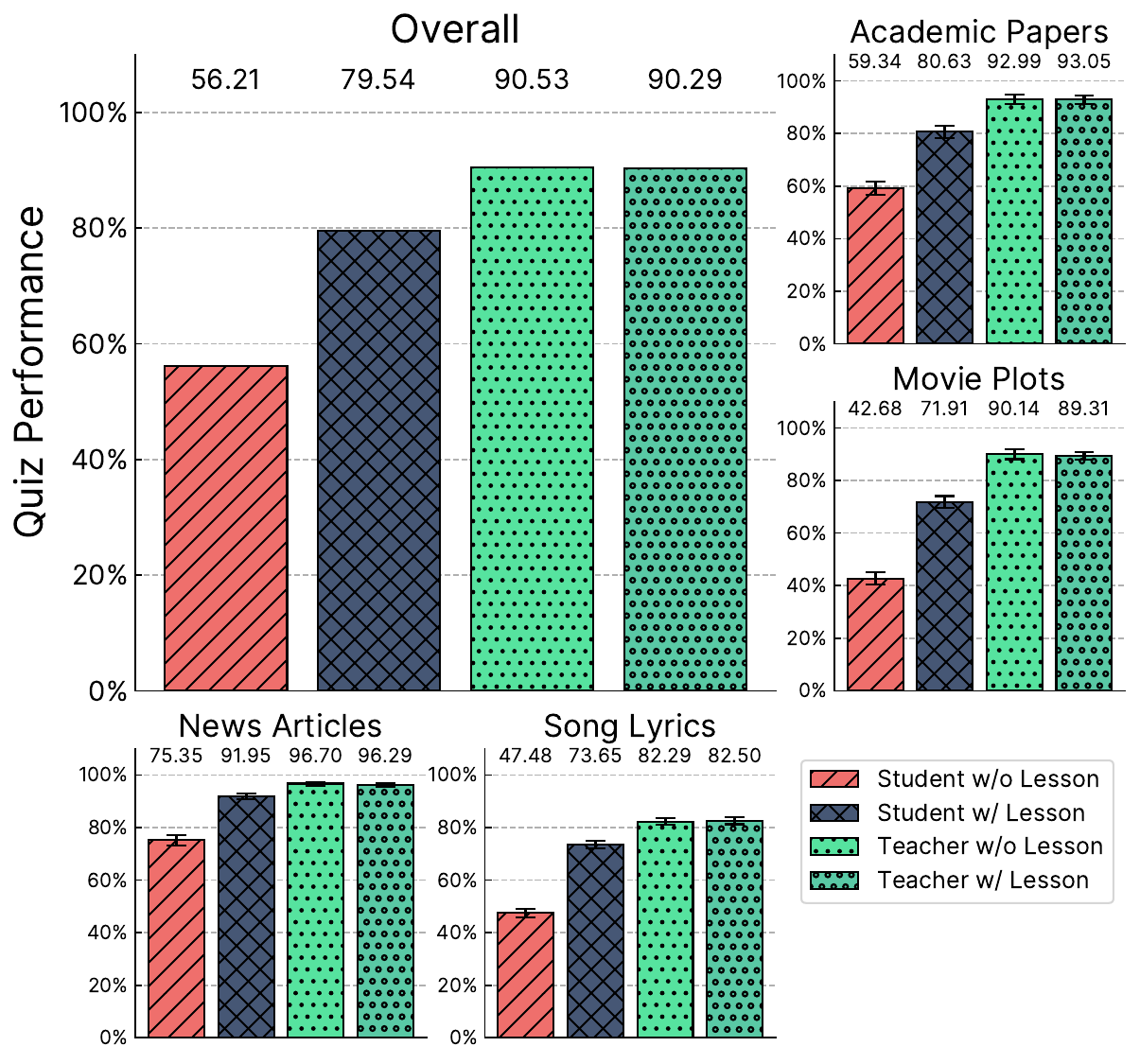} 
        \caption{\bf \large \llamalarge}
        \label{fig:llama-70b-static}
    \end{subfigure}
    \begin{subfigure}[b]{0.45\textwidth}
        \centering
        \includegraphics[width=0.9\textwidth]{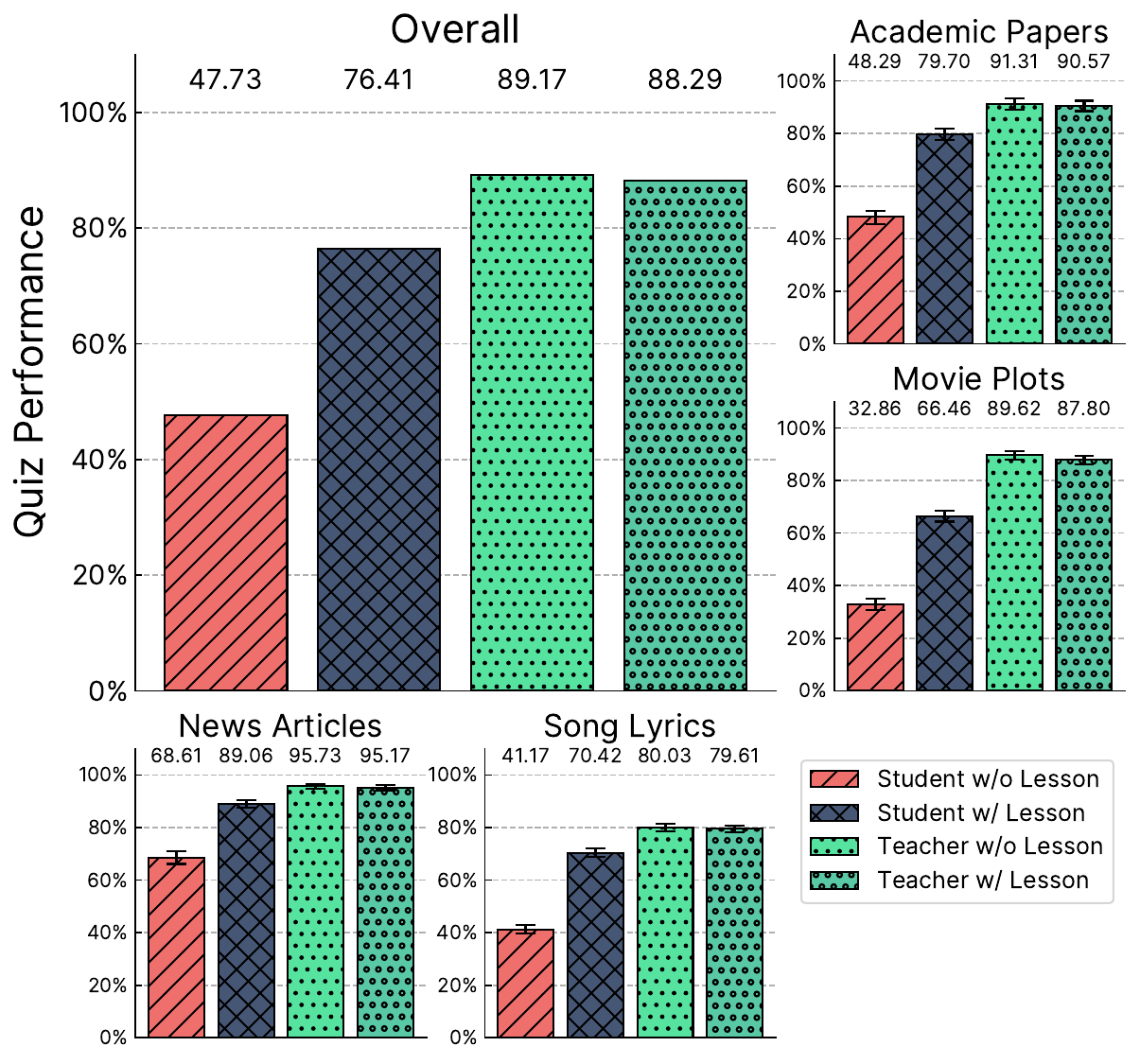} 
        \caption{\bf \large \gemmasmall}
        \label{fig:gemma-9b-static}
    \end{subfigure}
    \hfill
    \begin{subfigure}[b]{0.45\textwidth}
        \centering
        \includegraphics[width=0.9\textwidth]{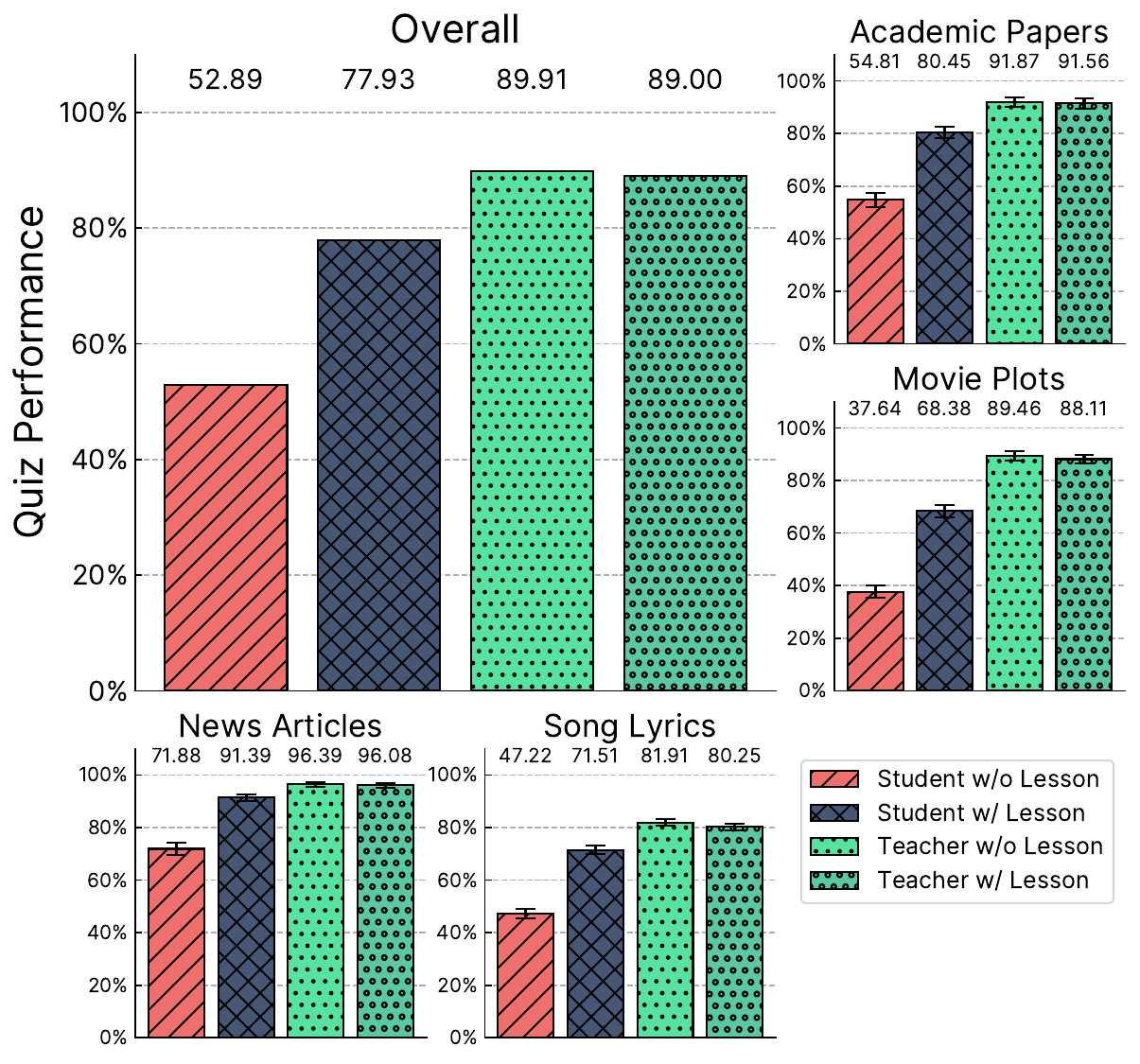} 
        \caption{\bf \large \gemmalarge}
        \label{fig:gemma-27b-static}
    \end{subfigure}
    
    \begin{subfigure}[b]{0.45\textwidth}
        \centering
        \includegraphics[width=0.9\textwidth]{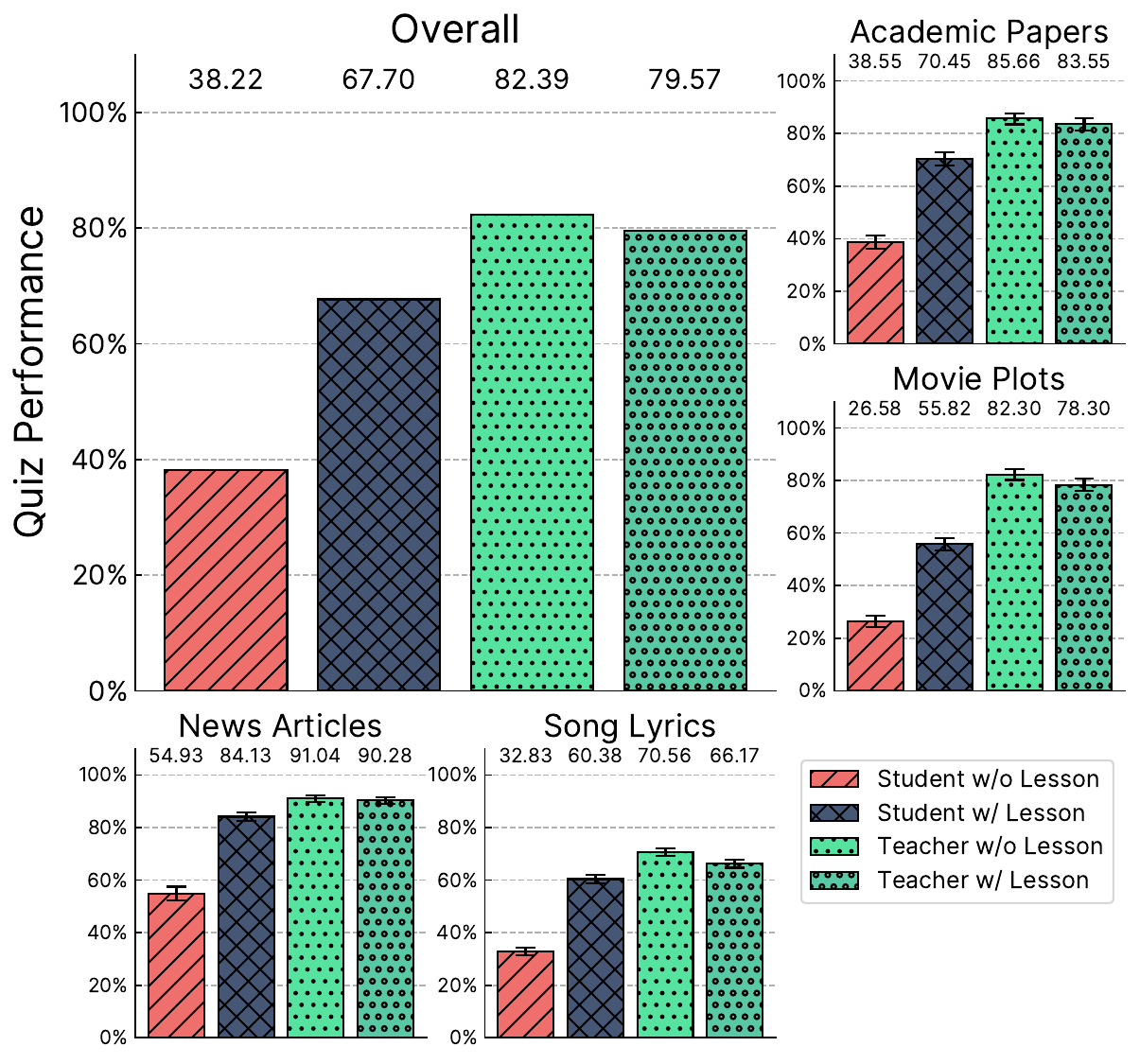} 
        \caption{\bf \large \ministral}
        \label{fig:ministral-8b-static}
    \end{subfigure}
    \hfill
    \begin{subfigure}[b]{0.45\textwidth}
        \centering
        \includegraphics[width=0.9\textwidth]{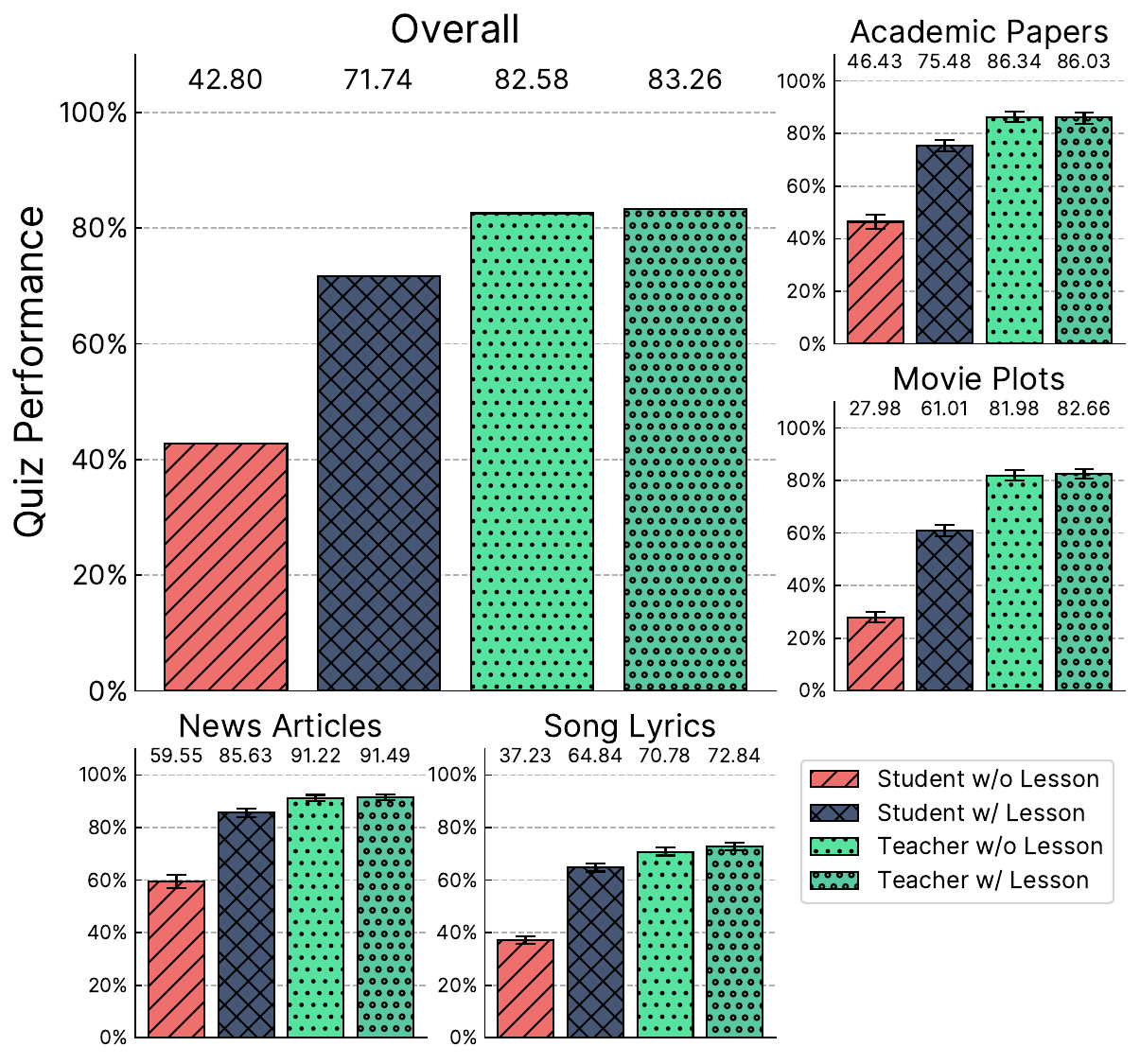} 
        \caption{\bf \large \mistralnemo}
        \label{fig:mistral-nemo-static}
    \end{subfigure}
    \begin{subfigure}[b]{0.45\textwidth}
        \centering
        \includegraphics[width=0.5\textwidth]{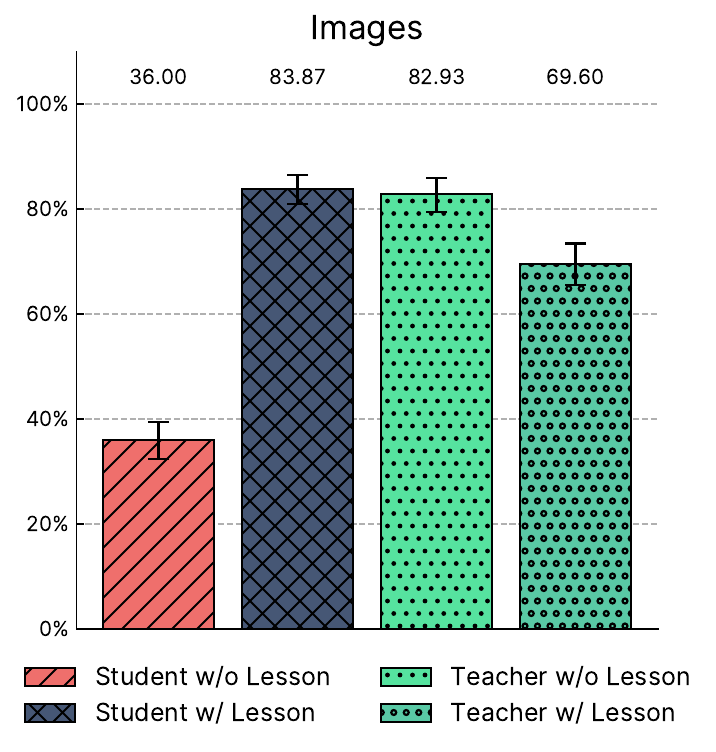} 
        \caption{\bf \large \llamav}
        \label{fig:llama-11b-static}
    \end{subfigure}
    \hfill
    \begin{subfigure}[b]{0.45\textwidth}
        \centering
        \includegraphics[width=0.5\textwidth]{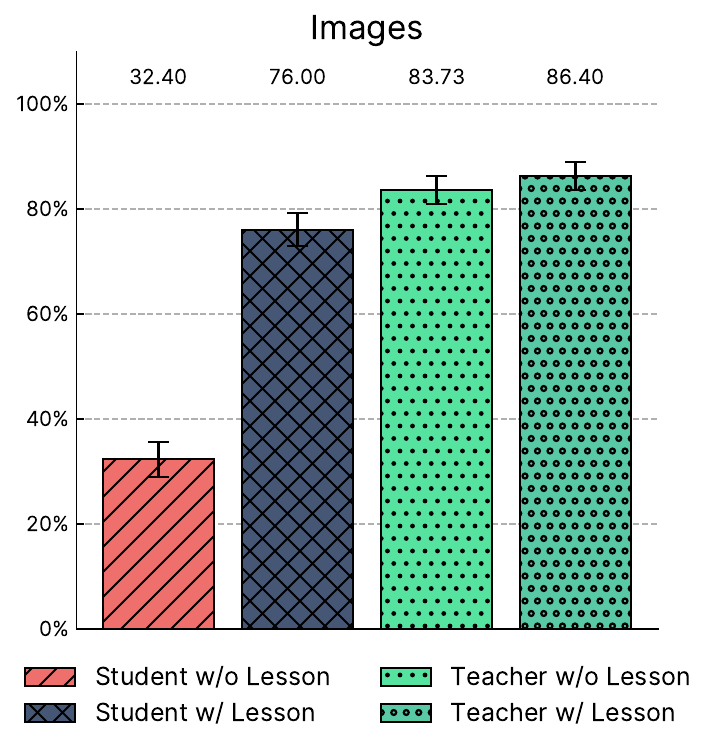} 
        \caption{\bf \large \pixtral}
        \label{fig:pixtral-static}
    \end{subfigure}
    \caption{Average quiz performance of student and teacher LLMs across different domains. Errorbars indicate the 95\% confidence interval calculated by bootstrap.}
    \label{fig:other_llms_static}
\end{figure*}

Figure \ref{fig:other_llms_static} shows the performance of various other LLMs on the static interactions. Teacher LLMs consistently outperform student LLMs, with their direct access to original material providing them with comprehensive contextual knowledge. Their near-perfect scores set a high bar for student LLMs. When teacher LLMs receive additional lessons, their performance improves only marginally. This suggests that, while summaries are beneficial, the original material already covers the essential information comprehensively, and teacher LLMs' access to detailed source material is crucial to their high performance. Overall, the substantial underperformance of the student LLM compared to the teachers highlights the challenge posed by our datasets to LLMs, leaving room for effective guidance by teachers.

\subsection{Additional Dynamic Results}

\begin{figure*}
    \centering
    \begin{subfigure}[b]{0.8\textwidth}
        \centering
        \includegraphics[width=0.9\textwidth]{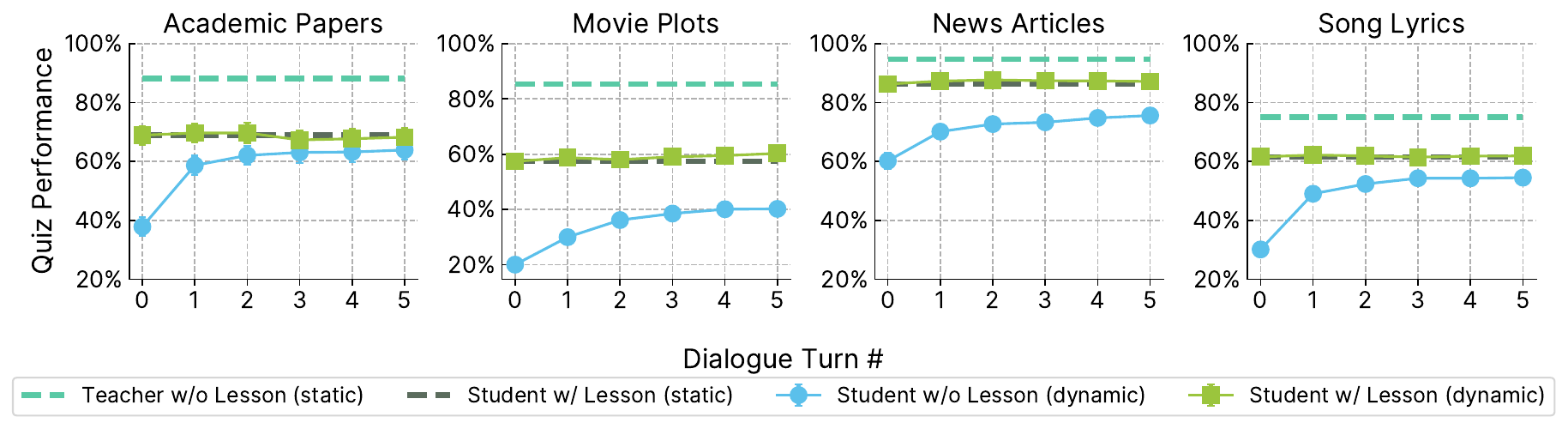} 
        \caption{\bf \large \llamasmall}
        \label{fig:llama-8b-dynamic}
    \end{subfigure}
    \hfill
    \begin{subfigure}[b]{0.8\textwidth}
        \centering
        \includegraphics[width=0.9\textwidth]{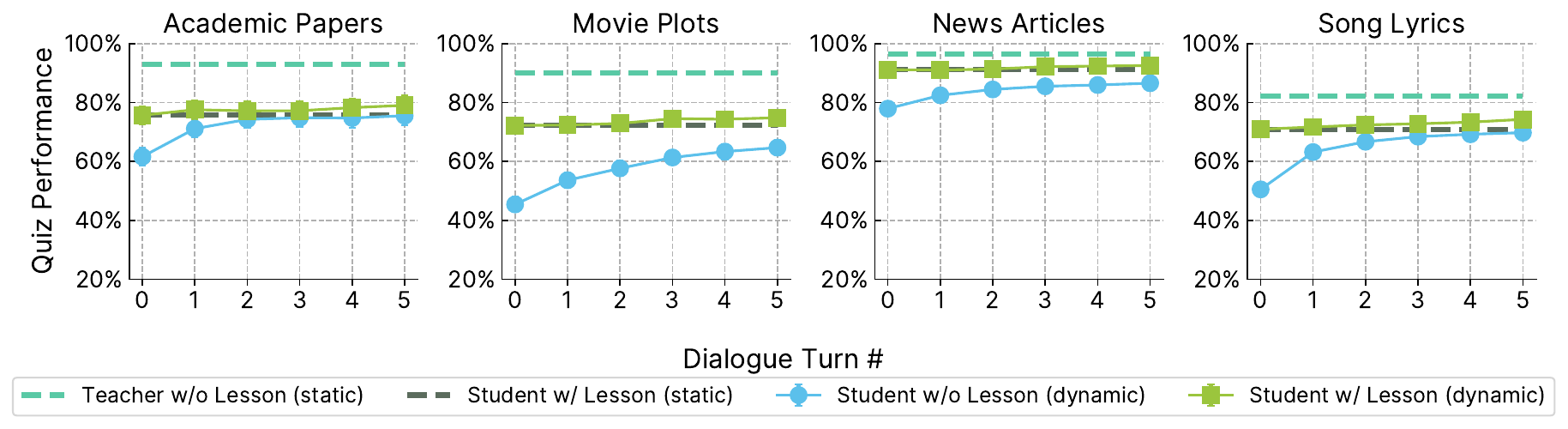} 
        \caption{\bf \large \llamalarge}
        \label{fig:llama-70b-dynamic}
    \end{subfigure}
    \begin{subfigure}[b]{0.8\textwidth}
        \centering
        \includegraphics[width=0.9\textwidth]{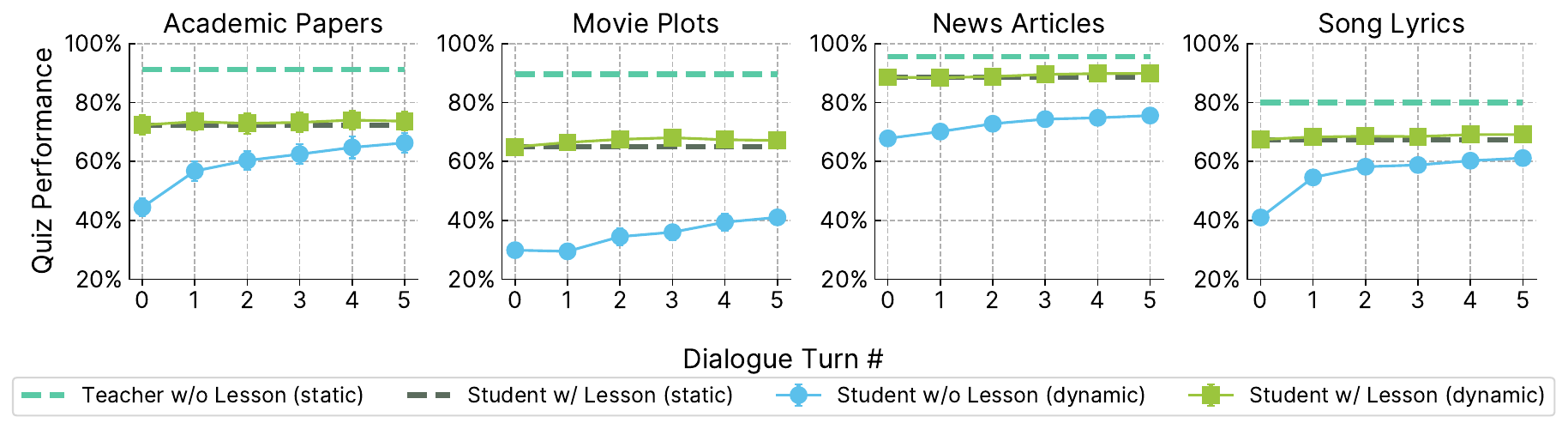} 
        \caption{\bf \large \gemmasmall}
        \label{fig:gemma-9b-dynamic}
    \end{subfigure}
    \hfill
    \begin{subfigure}[b]{0.8\textwidth}
        \centering
        \includegraphics[width=0.9\textwidth]{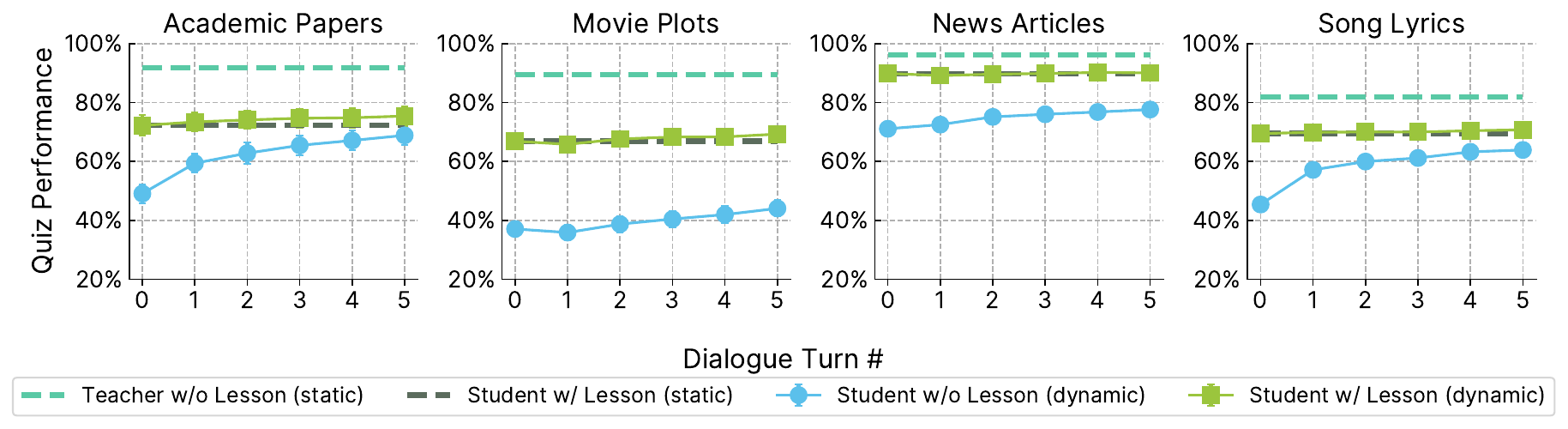} 
        \caption{\bf \large \gemmalarge}
        \label{fig:gemma-27b-dynamic}
    \end{subfigure}
    
    \begin{subfigure}[b]{0.8\textwidth}
        \centering
        \includegraphics[width=0.9\textwidth]{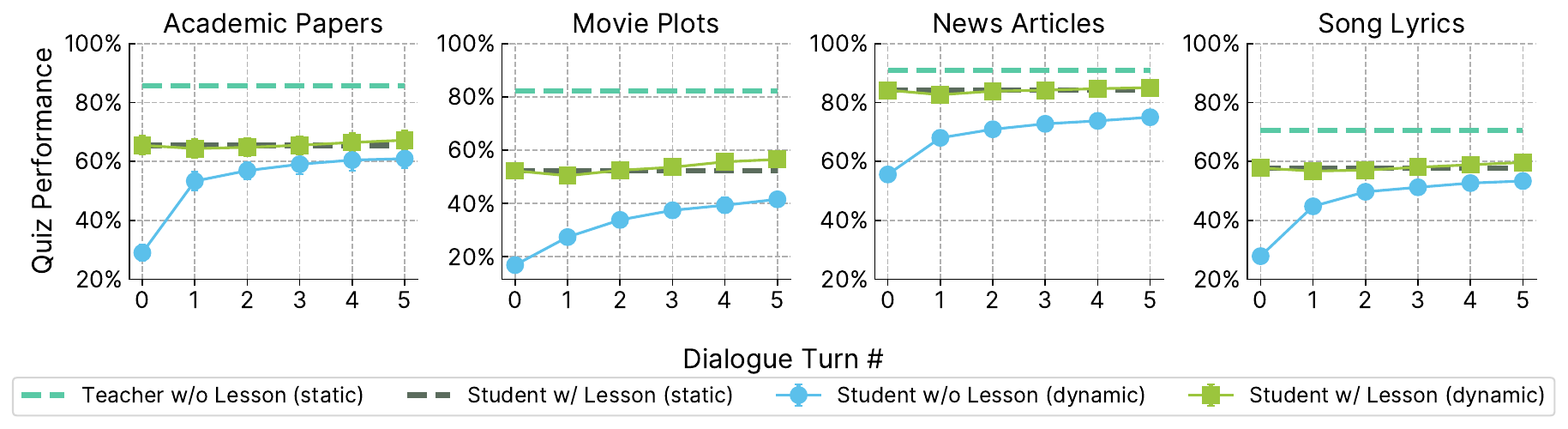} 
        \caption{\bf \large \ministral}
        \label{fig:ministral-8b-dynamic}
    \end{subfigure}
    \hfill
    \begin{subfigure}[b]{0.8\textwidth}
        \centering
        \includegraphics[width=0.9\textwidth]{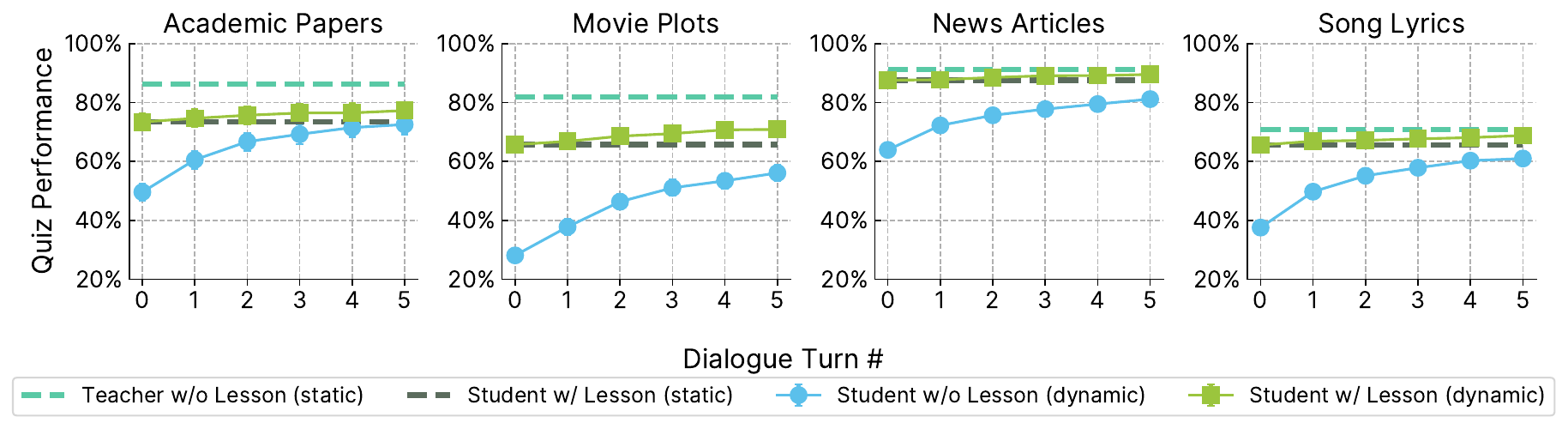} 
        \caption{\bf \large \mistralnemo}
        \label{fig:mistral-nemo-dynamic}
    \end{subfigure}
    \caption{Performance of student LLMs across various static and dynamic evaluation settings. Errorbars indicate the 95\% confidence interval calculated by bootstrap.}
    \label{fig:other_llms_dynamic}
\end{figure*}

Figure \ref{fig:other_llms_dynamic} shows the performance of various other LLMs on the dynamics interactions. Table \ref{tab:sampled_questions} provides some example questions the \texttt{gpt-4o} student LLM asked within dynamic interactions.

\begin{figure*}
    \begin{subfigure}[b]{0.3\textwidth}
        \centering
        \includegraphics[width=0.9\textwidth]{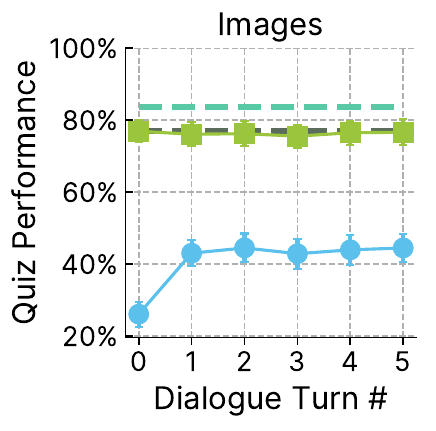} 
        \caption{\bf \large \pixtral}
        \label{fig:pixtral-dynamic}
    \end{subfigure}
    \hfill
    \begin{subfigure}[b]{0.6\textwidth}
        \centering
        \includegraphics[width=0.9\textwidth]{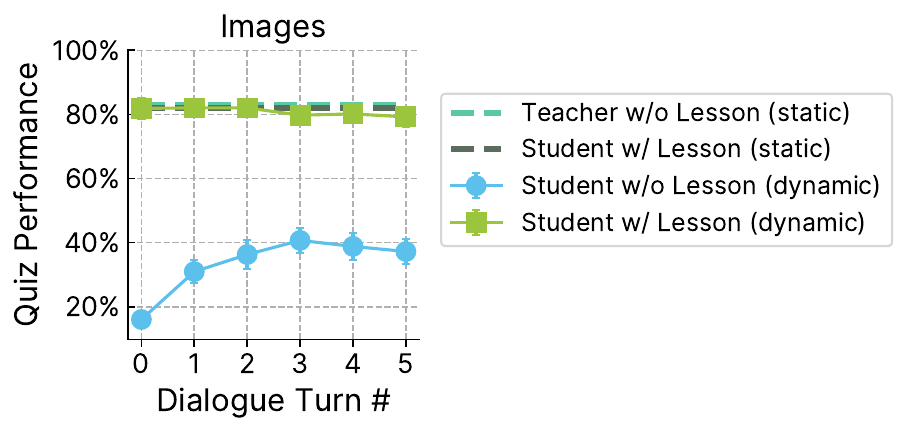} 
        \caption{\bf \large \llamav}
        \label{fig:llamav-dynamic}
    \end{subfigure}
    \caption{Performance of student MLLMs across various static and dynamic evaluation settings. Errorbars indicate the 95\% confidence interval calculated by bootstrap.}
    \label{fig:other_vllms_dynamic}
\end{figure*}

\begin{figure*}
    \centering
    \includegraphics[width=0.9\textwidth]{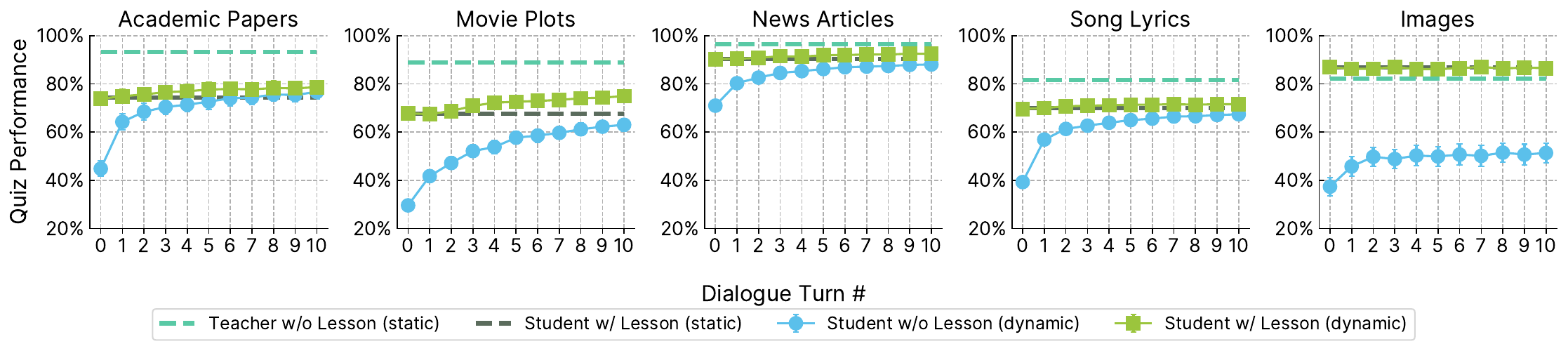}
    \caption{Performance of \gptmini across various static and dynamic evaluation settings when provided with more interaction rounds. Errorbars indicate the 95\% confidence interval calculated by bootstrap.}
    \label{fig:other_vllms_dynamic_longer}
\end{figure*}

\onecolumn
\begin{longtable}{@{}lll p{8cm}@{}}
\caption{Sampled Student Questions from Interactions} \label{tab:sampled_questions} \\
\toprule
\textbf{Model} & \textbf{Domain} & \textbf{Round} & \textbf{Question} \\
\midrule
\endfirsthead

\multicolumn{4}{c}%
{{\bfseries \tablename\ \thetable{} -- continued from previous page}} \\
\toprule
\textbf{Model} & \textbf{Domain} & \textbf{Round} & \textbf{Question} \\
\midrule
\endhead

\midrule \multicolumn{4}{r}{{Continued on next page}} \\
\endfoot

\bottomrule
\endlastfoot

\multirow{12}{*}{\gptmini}
 & news articles & round 4 & What health or safety considerations should individuals keep  in mind when shaping their eyebrows, especially when  adopting trends like thin brows? \\
 & news articles & round 3 & What specific actions or policies have been proposed or  implemented by colleges in response to the House committee's  findings on antisemitic harassment? \\
 & movie plots & round 1 & What is the central theme explored in "The Bikeriders," and  how does it reflect the culture or lifestyle of biker  communities? \\
 & movie plots & round 5 & What role does humor play in movies featuring "Space Cadet"  characters, and how does it enhance the storytelling or  character development? \\
 & movie plots & round 3 & What themes and motifs are commonly associated with hit man  characters in films, and how do these elements enhance the  storytelling? \\
 & academic papers & round 2 & How do Convolutional Neural Networks (CNNs) process seismic  data differently than traditional data analysis methods? \\
 & academic papers & round 3 & What specific practical applications or investment  strategies can be developed using "Dynamic Factor Allocation  Leveraging Regime-Switching Signals"? \\
 & academic papers & round 3 & What were the key findings of the study regarding the  relationship between the shape of brain connections and  cognitive performance? \\
 & song lyrics & round 1 & What inspired Kylie Minogue to create "Hold On To Now," and  what themes does the song explore? \\
 & song lyrics & round 5 & How does the title "Comatose" relate to the overall themes  presented in the song, and what significance does it hold in  the context of the lyrics? \\
\multirow{12}{*}{\llamasmall} & news articles & round 5 & How does the role of social media platforms, particularly in  amplifying and spreading anti-vaccine misinformation,  contribute to the decline in public confidence in vaccines,  as seen during the Trump presidency? \\
 & news articles & round 2 & What motivated Nima Rinji Sherpa to pursue climbing the  "eight-thousanders" despite being just 17 years old, and how  has he maintained his interest and commitment to  mountaineering throughout his journey? \\
 & news articles & round 5 & What specific actions or responses from others, if any,  demonstrate the impact and effectiveness of D'Pharaoh Woon-  A-Tai's bold Emmys statement in raising awareness about the  Missing and Murdered Indigenous Women movement? \\
 & movie plots & round 3 & Considering the narrative's creative interpretation of  Japanese folklore, how does the character of Hiiragi's  transformation from a suppressed emotional state to becoming  an Oni relate to the traditional Japanese concept of "Yūgen"  (mysterious and profound sense of the beauty of the world),  and what insights does it offer into the human experience? \\
 & movie plots & round 2 & What specific historical events or cultural movements might  have influenced the meaning of "The Beautiful Game" to shift  from a focus solely on the sport's elegance and skill to its  positive social impact and role in promoting social change? \\
 & movie plots & round 2 & How does the concept of deception, such as Marisol's attempt  to hide her husband's condition, relate to the broader theme  of social dynamics during a zombie apocalypse, and what  implications does it have for the characters and the story? \\
 & academic papers & round 4 & What specific contributions have Deborah Colleen and Steve  Kol steril fl iterators Arthur Choi have made in the  research of "Superpolynomials of Algebraic Links", and how  have their contributions impacted the broader field of  mathematics? \\
 & academic papers & round 3 & What are the typical CT scan characteristics or image  features that facilitate the employment of AI algorithms,  such as CNNs, in identifying and distinguishing between  malignant and benign pulmonary nodules? \\
 & song lyrics & round 3 & What specific line or lyrics in the song "Find Myself Again"  by Blanks is an example of expressing the tension between  the desire for comfort and security, and the need for growth  and change, as mentioned in the teacher's current  understanding? \\
\multirow{12}{*}{\llamalarge} & news articles & round 5 & What are some potential long-term consequences for high-  income families making over \$150,000 who are  living paycheck to paycheck, and how might this impact their  financial stability and overall well-being in the future? \\
 & news articles & round 3 & How do the different portrayals of Joan of Arc in art,  fashion, and film reflect and shape societal attitudes  towards women, power, and identity, and what do these varied  interpretations reveal about the cultural significance of  her legend? \\
 & movie plots & round 3 & How does the historical context of the 1924 Paris Olympics  and the societal attitudes towards women in sports during  that era influence the narrative and themes presented in  "Young Woman and the Sea"? \\
 & movie plots & round 5 & What is the significance of the Cheeks family circus in  "Saving Bikini Bottom: The Sandy Cheeks Movie", and how does  its final performance at the Krusty Krab relate to the  overall themes of "Family and Support" and "The Power of  Friendship"? \\
 & movie plots & round 1 & Can you explain the core theme or message that the concept  of "Upgraded" typically conveys in movie plots, and how it  relates to human emotions or societal commentary? \\
 & academic papers & round 4 & How do researchers using algebraic approaches to  cosmological integrals ensure the accuracy and reliability  of their computations and results, especially when dealing  with complex and multivariable systems? \\
 & academic papers & round 2 & Can you explain how the integration of convolutional neural  networks with Wiener filters addresses the limitations of  traditional methods in characterizing seismic isolation  systems, particularly in the context of gravitational wave  detectors? \\
 & academic papers & round 5 & How do the results from coarse-grained modeling,  specifically those related to gas adsorption at the  nanoparticle level, contribute to the development of more  efficient and selective metal-organic framework/polymer  composite materials for gas separation applications? \\
 & song lyrics & round 2 & What role does the artists' use of language play in  conveying the themes and emotions present in the song  "Muppy", and how does their lyrical style contribute to the  overall tone and message of the song? \\
 & song lyrics & round 2 & How does the artist, Quando Rondo, use the themes of loss  and heartbreak to underscore the struggle with the  consequences of one's actions in the song "Gotta Do Better  \& Pray"? \\
\multirow{12}{*}{\ministral} & news articles & round 1 & 1. What context or background information led to the  discovery of a long-lost copy of the US Constitution in a  filing cabinet? \\
 & news articles & round 3 & What emotional or musical characteristics were notable in  the recently discovered Chopin waltz to distinguish it from  other works of his? \\
 & news articles & round 3 & What are the key legal arguments the district attorney has  used in his previous clemency requests in other cases, if  any? \\
 & movie plots & round 5 & What role does comedy play in the story arc of "The People's  Joker" in relation to the protagonist's personal journey and  character development? \\
 & movie plots & round 1 & What is the primary theme of the movie "Road House"? \\
 & movie plots & round 2 & What are the key symbols and imagery used in "Demon Slayer:  Kimetsu no Yaiba – To the Hashira Training"? \\
 & academic papers & round 3 & Could you explain the role of geography in this concept, and  specifically how the Soviet Union's geographical interests  and strategic positions influenced the propaganda's  representation of cities?  *Teacher, please respond before I  ask the next question.* \\
 & academic papers & round 2 & What specific challenges does the survey identify as  prominent in the application of large language models to  algorithm design? \\
 & academic papers & round 1 & Which are the primary challenges addressed by the concept of  “Privacy for Free in the Over-Parameterized Regime”? \\
 & song lyrics & round 5 & Why does the song "No Love" by Marjinal use Whatever  attendance worksheet backwards by rachal owner lapie ? \\
 & song lyrics & round 1 & What are the primary themes and emotions conveyed in the  song "Darling" by Ali Gatie? \\
 & song lyrics & round 1 & What are the primary themes present in the song "Or What" by  NLE Choppa \& 41? \\
\multirow{12}{*}{\mistralnemo} & news articles & round 2 & What was the reason given by Iranian authorities for  executing Jamshid Sharmahd? \\
 & news articles & round 2 & What did the DNA analysis specifically reveal about the fate  of the crew of Sir John Franklin's expedition? \\
 & news articles & round 4 & What are some unique interactive experiences or activities  visitors can expect at this barbecue museum, besides  learning about different styles and regions? \\
 & movie plots & round 5 & Can you analyze the role of trust in the relationship  between the two "Wolfs" or fixers in the movie plots domain?  How does their mutual trust affect their ability to work  together and maintain their loyalty and honor? \\
 & movie plots & round 5 & Can you provide an example of a movie where "It's What's  Inside" is explored in a non-fantastical or supernatural  way, using a different plot device? \\
 & movie plots & round 1 & What is the origin story of Ricky Stanicky in the movie plot  domain? \\
 & academic papers & round 4 & How does the presence of alternative folds affect  AlphaFold's ability to predict the functional properties of  proteins, such as their interactions with other biomolecules  or their cellular localization? \\
 & academic papers & round 5 & What are the potential benefits of using a light-weight  feature-attention module in the FACMIC model, aside from  reducing communication costs? \\
 & academic papers & round 2 & How does the brain-aware readout layer improve  interpretability compared to traditional approaches? \\
 & song lyrics & round 4 & What are the specific themes and messages conveyed in "Stuck  Inside (CG5 Remix)" and how does the collaboration between  Black Gryph0n, CG5, \& The Living Tombstone  enhance or alter these themes? \\
 & song lyrics & round 0 & N/A \\
\multirow{12}{*}{\gemmasmall} & news articles & round 1 & What is the geographic location of this shipyard and the  potential impact of the fire on the surrounding environment? \\
 & news articles & round 3 & What are some historical examples of unexpected events that  significantly impacted the jobs report and how did they  influence the economic outlook at the time? \\
 & movie plots & round 5 & What are some specific  scenes or moments from "Problemista"  that exemplify Alejandro's unwavering determination in the  face of adversity? \\
 & movie plots & round 3 & Could you give some examples of movies that depict civil  wars and how they exemplify those themes of loyalty and  betrayal you mentioned? \\
 & movie plots & round 5 & What are some concrete examples of "societal control" and  "individual autonomy"  struggles depicted in movies that  explore the concept of "Humane"? \\
 & academic papers & round 3 & What are some of the specific advantages of using  "Fluorescence mass spectrometry" for detecting trapped O+2  ions compared to other detection methods? \\
 & academic papers & round 3 & Could you provide an example of a real-world application  where the Improved Velocity-Verlet Algorithm proves  particularly beneficial for the Discrete Element Method? \\
 & academic papers & round 5 & What specific performance metrics were used to evaluate the  effectiveness of EViT-UNet for medical image segmentation,  and how did it compare to existing methods in terms of  accuracy, speed, and resource usage? \\
 & song lyrics & round 5 & How does the call-and-response element of the song  contribute to the theme of unity and togetherness? \\
 & song lyrics & round 2 & How does DRAM use imagery and metaphors in "Marry Me" to  convey the depth and intensity of his feelings? \\
 & song lyrics & round 3 & Can you tell me more about the musical structure of "Back To  The Wall" and how it contributes to the overall message of  the song? \\
\multirow{12}{*}{\gemmalarge} & news articles & round 1 & What specific historic milestone did Nikola Jokic achieve? \\
 & news articles & round 5 & How do the traditional Ayurvedic and homeopathic practices  offered at Soukya International Holistic Health Centre align  with King Charles' known interest in alternative medicine  and sustainability? \\
 & movie plots & round 5 & How does the film "Apartment 7A" use sound and music to  create an atmosphere of suspense and unease? \\
 & movie plots & round 1 & What are some common themes or motifs explored in stories  within the "My Oni Girl" genre? \\
 & academic papers & round 3 & What are some examples of specific applications where the  Diffusion-PINN Sampler has been successfully used or is  particularly well-suited? \\
 & academic papers & round 5 & What are some potential limitations or challenges associated  with this method of modeling dynamic neural activity? \\
 & academic papers & round 3 & What are the potential benefits and implications of using  Large Language Models for algorithm design outlined in the  survey? \\
 & song lyrics & round 4 & Can you provide any insights into the lyrical choices made  by American Authors in their rendition of "Sweet Caroline"  compared to Neil Diamond's original version? \\
 & song lyrics & round 2 & Can you provide examples of specific lyrics from the song  that illustrate the themes of neighborhood violence and  retaliation? \\
\end{longtable}
\onecolumn
\twocolumn
\pagebreak

\section{Dataset Creation}
\label{app:dataset_creation}
\subsection{Data Collection}

Datasets were compiled by using a mix of API, scraping, and manual collection of data starting from January 2024 to obtain song lyrics, movie plots, news articles, and academic papers.\footnote{This temporal criterion was strategically chosen to ensure that the data used was not previously encountered by the GPT-4o model, thus eliminating potential biases or prior knowledge that could influence the model's performance in teaching and learning scenarios.} In addition to textual data, the Visual Question Answering (VQA) dataset from the COCO image collection was utilized to add a multimodal dimension to the context preparation, further challenging the instructional capabilities of the models under study. Table \ref{tab:dataset-composition} shows a breakdown of the dataset distribution, showcasing the diversity and scope of the data collected.

\begin{table}[t]
\centering
\begin{tabular}{@{}lrr@{}}
\toprule
\textbf{Domain Name} & \begin{tabular}{r@{}}\textbf{Context}\\ \textbf{Count}\end{tabular} & \begin{tabular}{r@{}}\textbf{Avg. \#}\\\textbf{Tokens}\end{tabular} \\
\midrule
Images & 150 & - \\
Movie Plots & 214 & 671.3 \\ 
Song Lyrics & 467 & 296.9 \\ 
Academic Papers & 170 & 1560.8 \\ 
\quad Computer Science & 23 & 1438.5 \\ 
\quad Economics & 16 & 1424.2 \\ 
\quad Electrical Engineering & 25 & 1443.3 \\ 
\quad Mathematics & 23 & 1667.6 \\ 
\quad Physics & 22 & 1781.7 \\ 
\quad Quantitative Biology & 22 & 1752.1 \\ 
\quad Quantitative Finance & 15 & 1462.4 \\ 
\quad Statistics & 24 & 1507.1 \\ 
News Articles & 346 & 1163.7 \\ 
\quad Business & 38 & 1068.5 \\ 
\quad Entertainment & 30 & 627.7 \\
\quad Health & 28 & 1372.8 \\ 
\quad Politics & 43 & 1690.6 \\ 
\quad Science & 27 & 1406.4 \\ 
\quad Sports & 41 & 1232.1 \\ 
\quad Style & 51 & 1023.0 \\ 
\quad Travel & 10 & 1026.6 \\ 
\quad US News & 28 & 1166.7 \\ 
\quad World News & 50 & 981.7 \\ 
\midrule
\textbf{Total} & 1,347 & 967.1 \\ 
\bottomrule
\end{tabular}
\caption{Dataset composition including context counts across domains and average token counts.}
\label{tab:dataset-composition}
\end{table}

\begin{table*}[htbp]
\centering
\scriptsize
\caption{Feature Names, Descriptions, and Details}
\label{tab:features_regression}
\begin{tabular}{@{}p{3cm} p{5cm} p{7cm}@{}}
\toprule
\textbf{Feature Name} & \textbf{Description} & \textbf{Details} \\
\midrule
\multicolumn{3}{l}{\textbf{Question-Level Features}} \\
\midrule
\textbf{Question Length} & Number of tokens or words in the student's question. & Token count via spaCy. \\
\textbf{Question Complexity} & Syntactic complexity via average parse tree depth. & Calculated using spaCy's dependency parsing. \\
\textbf{Lexical Sophistication} & Average word length as a proxy for rarity or frequency. & Based on average token length. \\
\textbf{Named Entity Count} & Number of named entities in the student's question. & Utilizes spaCy's Named Entity Recognition (NER). \\
\textbf{Question Informativeness} & Number of unique domain-relevant keywords present. & Intersection with predefined domain keywords. \\
\textbf{Question Directness} & Presence of a question mark indicating clarity. & Binary: 1 if '?', else 0. \\
\textbf{Politeness/Hedging} & Count of politeness or hedging words like "maybe" or "could". & Uses a predefined set of hedging words. \\
\textbf{Question Type} & Categorizes into types such as Who, What, Where, etc. & Binary indicator for specific question starters. \\
\textbf{Question Novelty} & Semantic difference from previous questions. & Cosine similarity of embeddings (computed with a hash function). \\
\textbf{Question Specificity} & Focus based on named entities presence. & Binary: 1 if entities >0, else 0. \\
\textbf{Bloom's Taxonomy} & Corresponding Bloom's taxonomy level \citep{Bloom1966TaxonomyOE}. &  A binary feature for each level. Utilizes \texttt{gpt-4o} to decide the taxonomy. \\

\midrule
\multicolumn{3}{l}{\textbf{Teacher-Response-Level Features}} \\
\midrule
\textbf{Response Length} & Number of tokens or words in the teacher's response. & Token count via spaCy. \\
\textbf{Info Density} & Ratio of informational tokens (nouns, proper nouns) to total tokens. & Based on POS tagging with spaCy. \\
\textbf{Response Novelty} & Amount of new content vs. previous responses. & Cosine similarity of embeddings (computed with a hash function). \\
\textbf{Response Correctness} & Factual correctness via QA model score. & Utilizes Hugging Face's QA pipeline (distilbert-base-cased-distilled-squad). \\
\textbf{Response Completeness} & Whether the response fully addresses the question. & Binary based on QA model score (>0.5 considered complete). \\
\textbf{Response Complexity} & Syntactic complexity of the teacher's response. & Calculated using spaCy's dependency parsing. \\
\textbf{Response Sentiment} & Sentiment score for the response. & Utilizes Hugging Face's text classification pipeline (clapAI/roberta-base-multilingual-sentiment) \\
\textbf{Entity Diversity} & Variety of named entities in the response. & Utilizes spaCy's NER to extract and count unique entities. \\
\textbf{Temporal Positioning} & Presence of chronological cues in the response. & Count of temporal keywords. \\
\textbf{Use of Examples} & Presence of illustrative phrases like "for example". & Binary based on phrases like "for example", "such as". \\

\midrule
\multicolumn{3}{l}{\textbf{Interaction-Dynamics Features}} \\
\midrule
\textbf{Turn Index} & Current round number in the interaction. & Sequential indexing starting from 1. \\
\textbf{Cumulative Exposure} & Number of unique facts introduced so far. & Unique token count. \\
\textbf{Student Adaptation} & Change in question complexity from the previous round. & Difference in complexity scores. \\
\textbf{Teacher Adaptation} & Change in response complexity from the previous round. & Difference in complexity scores. \\
\textbf{Information Gain} & Semantic difference from the previous response. & Cosine similarity of embeddings (e.g., SentenceTransformers). \\
\textbf{Topic Shifts} & Shift to a new aspect of the concept. & Cosine similarity of embeddings. \\
% \textbf{Q-A Alignment} & Alignment between question and response. & Cosine similarity of embeddings (e.g., SentenceTransformers). \\
\textbf{Unanswered Queries} & Count of unanswered questions. & Uses \texttt{gpt-4o} to determine whether a given response answers the corresponding question. \\
\textbf{Progressive Elaboration} & Degree of building upon earlier knowledge. & Based on response length trends. \\
\textbf{Student Context Coverage} & The cumulative overlap between question entities so far and context entities. & Utilizes spaCy's NER to extract and count unique entities. \\
\textbf{Teacher Context Coverage} & The cumulative overlap between response entities so far and context entities. & Utilizes spaCy's NER to extract and count unique entities. \\
\textbf{Student Quiz Coverage} & The cumulative overlap between question tokens so far and quiz question tokens. & The ratio of unique quiz tokens covered by the questions. \\
\textbf{Teacher Quiz Coverage} & The cumulative overlap between response tokens so far and quiz question tokens. & The ratio of unique quiz tokens covered by the responses. \\
\textbf{Student Semantic Alignment} & The average maximum similarity between each quiz question and the questions asked so far. & Cosine similarity of embeddings. \\
\textbf{Teacher Semantic Alignment} & The average maximum similarity between each quiz question and the responses so far. & Cosine similarity of embeddings. \\
\midrule
\multicolumn{3}{l}{\textbf{Linguistic/Style Features}} \\
\midrule
\textbf{Lexical Diversity (Student)} & Type-token ratio in student questions. & Unique/total words ratio. \\
\textbf{Lexical Diversity (Teacher)} & Type-token ratio in teacher responses. & Unique/total words ratio. \\
\textbf{Domain-Specific Terms} & Frequency of domain keywords in text. & Intersection with predefined domain keywords. \\
\textbf{Sentence Length Variability} & Std. deviation of sentence lengths in responses. & Calculated using spaCy's sentence segmentation. \\
\textbf{Readability Score} & Readability via Flesch-Kincaid or similar. & Utilizes the \texttt{textstat} library. \\
\textbf{Passive Voice Count} & Number of passive constructions in the response. & spaCy dependency \texttt{auxpass} count. \\
\textbf{Modal Language Count} & Frequency of modal/uncertain words. & Based on a predefined set of modal words. \\

\midrule
\multicolumn{3}{l}{\textbf{Semantic/NLP Features}} \\
\midrule
\textbf{Semantic Similarity to Summary} & Alignment with reference summary. & Cosine similarity of embeddings (computed with a hash function). \\
\textbf{Coreference Complexity} & Number of coreference chains. & Uses spaCy-coref pipeline. \\
\textbf{Semantic Cohesion} & Similarity with all previous responses. & Cosine similarity of embeddings (computed with a hash function). \\
\textbf{Coverage of Key Plots} & Fraction of key plot elements mentioned. & Presence of key plot terms. \\

\midrule
\multicolumn{3}{l}{\textbf{Performance/Contextual Features}} \\
\midrule
\textbf{Prior Knowledge Estimate} & Initial knowledge based on pre-interaction quiz. & Initial quiz score. \\
\textbf{Student Confidence} & Quiz accuracy score (0-100). & Numeric value from \texttt{student\_evaluation}. \\
\textbf{Improvement in Questions} & Trend in question clarity/specificity. & Difference from first round complexity. \\
\textbf{Instruction-Following Score} & Adherence to teacher instructions. & Uses reward model scores as proxy. (Ray2333/GRM-Gemma2-2B-rewardmodel-ft) \\
\textbf{Redundancy in Answers} & Fraction of repeated information. & Token overlap with previous responses. \\
\textbf{Politeness/Social Cues} & Presence of courteous language. & Predefined polite words like "please", "thank you". \\
\textbf{Meta-Linguistic Feedback} & References to previous turns. & Binary: Phrases like "as mentioned". \\
\bottomrule
\end{tabular}
\end{table*}

\paragraph{Movie Plots}
The dataset for movie plots was compiled by scraping Wikipedia pages under the \href{https://creativecommons.org/licenses/by-sa/3.0/}{Creative Commons Attribution-ShareAlike 3.0 License}. This method complies with Wikipedia's \href{https://wikitech.wikimedia.org/wiki/Robot_policy}{robot policy}, ensuring ethical scraping practices. Only movie plots released from January 2024 onwards were included to ensure data relevance and alignment with GPT-4o's latest knowledge cutoff date of October 2023. The scraping process adhered strictly to Wikipedia's terms, ensuring attribution is maintained, and derivatives follow the same licensing requirements. This inclusion ensures the experimental results remain unaffected by prior knowledge encoded in GPT-4o’s training data.

\paragraph{Song Lyrics}
Song lyrics were collected using the \href{https://docs.genius.com/\#/getting-started-h1}{Genius API} via the \href{https://lyricsgenius.readthedocs.io/en/master/}{LyricsGenius Python client}. Previously, scraping was used, but this was transitioned to API usage to comply with Genius's \href{https://genius.com/static/terms}{Terms of Service}, which explicitly prohibit scraping while permitting data retrieval via their API. Only lyrics from songs released after January 2024 were included, ensuring alignment with GPT-4o's knowledge cutoff. This transition to API usage guarantees the dataset’s legality and ethical compliance, avoiding any terms-of-service violations while maintaining the integrity of the collected data.

\paragraph{News Articles}
News articles were collected from CNN during a one-day span in November 2024 by downloading raw HTML pages. Articles were categorized into topics such as politics, world, business, and entertainment. Only articles published from January 2024 onwards were included. CNN's \href{https://www.cnn.com/2014/01/17/cnn-info/interactive-legal/index.html}{Terms of Service} permit automated content retrieval for academic purposes, provided it does not manipulate page views or server traffic. This ensured the legality of this data collection process. Furthermore, the inclusion of recent articles minimizes the risk of duplicating pre-existing knowledge in GPT-4o, ensuring up-to-date and unbiased context for research purposes.

\paragraph{Academic Papers}
Academic papers were sourced from arXiv using their \href{https://info.arxiv.org/help/api/tou.html}{API}, in compliance with their Terms of Use, which allow retrieval and utilization of e-prints for research purposes. Only papers released in October 2024 were included, focusing on fields such as computer science, mathematics, and economics. The first 1,500 words of each paper were extracted using arXiv's beta \href{https://arxiv.org/}{HTML renderer}. In cases where the HTML renderer was unavailable, PDFs were processed instead. This method ensured textual consistency, avoided formatting issues, and complied with arXiv's API guidelines. By restricting papers to those published after January 2024, the dataset ensures it is free from pre-existing knowledge encoded in GPT-4o's training data.

\paragraph{Images}
The Visual Question Answering dataset was sourced from the \href{https://cocodataset.org/#home}{COCO image collection}, which is available under the \href{https://creativecommons.org/licenses/by/4.0/}{Creative Commons Attribution 4.0 License}. This dataset enhances the multimodal aspect of the study by integrating visual contexts alongside text. While the textual datasets were restricted to content published after January 2024 to avoid influencing GPT-4o with pre-existing knowledge, the inclusion of older images from the COCO dataset does not present the same risk. Images, unlike text, do not carry direct semantic content that could be memorized or specifically encoded in a language model’s training data. Therefore, the age of the images is inconsequential to GPT-4o's ability to analyze and interpret visual information. This distinction justifies the inclusion of older images to expand the scope of visual contexts without compromising the experimental results.

\subsection{Quiz Questions}
\label{tab:question-types}
\paragraph{Quiz Question Complexities.} To ensure comprehensive evaluation of the LLM's learning capabilities, quiz questions were designed across three levels of complexity: Middle-School, College, and Graduate. Each level progressively increases in difficulty and depth, as described below.

\begin{itemize}
    \item \textbf{Middle-School Level Questions:} These questions test foundational understanding by focusing on basic recall of facts, definitions, or direct observations. They are simple and factual, ensuring accessibility for beginners.
    \item \textbf{College Level Questions:} These questions assess intermediate conceptual understanding, requiring interpretation of logical relationships, main ideas, causes, effects, and motivations. They are more challenging and engage with the material at a deeper level.
    \item \textbf{Graduate Level Questions:} These questions evaluate in-depth analytical and critical thinking skills, focusing on symbolic or thematic interpretation, synthesis of ideas, and evaluation of broader themes or theories. They involve the highest complexity, demanding advanced problem-solving or theoretical applications.
\end{itemize}

\begin{table*}[h!]
\centering
\begin{tabularx}{\textwidth}{l l X}
\toprule
\textbf{Domain} & \textbf{Difficulty} & \textbf{Example Quiz Questions} \\
\midrule
\multirow{3}{*}{Academic Papers} 
    & Middle-School & What does the acronym dMRI stand for in the context of the study? \\
    & College       & According to the document, what is the conjecture regarding every ribbon knot? \\
    & Graduate      & What overarching theme does the paper suggest through the study of DAHA and motivic superpolynomials? \\
\midrule

Images & — & What color are the flowers in the right garden bed? \\
\midrule

\multirow{3}{*}{Movie Plots} 
    & Middle-School & Who is the demigoddess mentioned in "The Casagrandes Movie"? \\
    & College       & Why does Gary refuse money from Madison initially? \\
    & Graduate      & Analyze the symbolic significance of Mae's satellite decryption key within the narrative. \\
\midrule

\multirow{3}{*}{News Articles} 
    & Middle-School & Why was Ellie the Elephant created as the New York Liberty's mascot? \\
    & College       & How does the artwork "Comedian" draw parallels to Marcel Duchamp's urinal according to commentators? \\
    & Graduate      & Which of the following best analyzes the potential thematic motivations behind Bob Costas' decision to retire from baseball play-by-play commentary after 42 years? \\
\midrule

\multirow{3}{*}{Song Lyrics} 
    & Middle-School & What is happening when "the lights go off" according to the singer? \\
    & College       & What is the stylistic effect of repeating the chorus in the song lyrics? \\
    & Graduate      & From a psychological perspective, how can the metaphor "off switch" be interpreted in terms of defense mechanisms? \\
\bottomrule
\end{tabularx}
\caption{Example quiz questions for each domain with different difficulty levels.}
\label{tab:example_quiz_questions}
\end{table*}

Table \ref{tab:example_quiz_questions} provides examples of the three types of questions generated for the evaluation process, as well as question generated across multiple domains.

\paragraph{Adversarial Quiz Question Generation.}To ensure the robustness and relevance of quiz questions, an adversarial generation process was implemented. This aimed to eliminate questions that could be answered by models using pre-existing knowledge rather than learning exclusively from the provided context. Given that LLMs are trained on large-scale open web-text datasets \citep{roberts-etal-2020-much}, they often possess world knowledge that can lead to misleading evaluations if the concepts are not sufficiently novel.

We observed that some generated questions were too easy for the models to answer due to several factors. Prior knowledge of the concepts being tested often allowed models to answer questions without relying on the provided context, as these concepts were part of the models' pre-training data. In some cases, the material in the context was sequential, serving as a follow-up to previously established information, making it easy for models to infer the answers. Additionally, some questions lacked sufficient cognitive challenge, failing to effectively test the model's concept acquisition abilities.

To address these issues, we implemented an iterative adversarial filtering strategy. For each context, a set of nine questions was initially generated using the \texttt{gpt-4o-2024-08-06} model. Each question was then tested with a smaller model, \texttt{gpt-4o-mini}, to identify questions that could be answered correctly without context. Those questions were filtered out and regenerated, repeating the process until the smaller model consistently missed the answer. To prevent infinite regeneration, we set a maximum of five attempts. If, after five iterations, the smaller model still answered the question correctly, we settled on the most recently generated version of the question.

This iterative adversarial approach ensured that each retained question effectively tested the model's ability to learn new concepts from the provided context, reducing reliance on prior knowledge.

\paragraph{Quiz Question Validation.} To ensure the quality and relevance of the generated quiz questions, a thorough manual evaluation process was conducted. This process aimed to validate the effectiveness of the questions in assessing the concept-learning abilities of student LLMs across various textual domains, including News Articles, Academic Papers, Song Lyrics, and Movie Plots.

A carefully selected sample of at least 50 questions was evaluated for each domain, ensuring both broad and balanced coverage. For domains with subdomains (e.g., News Articles and Academic Papers), this involved choosing sets of three questions—one at each difficulty level (Middle School, College, and Graduate)—from each subdomain and repeating this process until the target sample size was reached. For instance, the News Articles domain, composed of 10 categories, was sampled in two rounds of selection (3 questions × 10 categories × 2 rounds = 60 questions), and Academic Papers, with 8 subfields, was sampled in three rounds (3 questions × 8 subfields × 3 rounds = 72 questions). For domains without subdomains, like Song Lyrics and Movie Plots, 51 questions were chosen through a similar iterative approach, randomly selecting triplets that included all three difficulty levels each time. This method ensured that the evaluation set was both representative of content diversity and reflective of the full spectrum of cognitive challenges the quiz was designed to assess.

For domains without subdomains, such as Song Lyrics and Movie Plots, questions were selected randomly but still included a balanced mix of difficulty levels. This ensured diversity in the evaluation set while maintaining alignment with the domain's unique characteristics.

The evaluation focused on three key criteria to determine question quality: (1) whether the question was suitable for testing student understanding, (2) whether it was answerable using the provided context alone, and (3) whether it avoided requiring knowledge beyond the provided context to grasp the concept. The results of this evaluation indicated that over 97\% of the reviewed questions satisfied all three criteria, demonstrating their effectiveness and relevance in the quiz phase. See Table \ref{tab:expanded_evaluation_summary} for the breakdown of the quiz validation results across each domain and corresponding question criteria.

This manual verification process, combined with adversarial filtering during question generation, ensured that the quiz questions were of high quality and closely aligned with the intended learning objectives for each domain.

\begin{table}[h!]
\centering
\small
\setlength{\tabcolsep}{4pt} % Reduce space between columns
\begin{tabular}{l r r}
\toprule
\textbf{Category} & \textbf{\# Evaluated} & \textbf{Yes (\%)} \\
\midrule
\multicolumn{3}{l}{\textbf{Question 1: Suitable for testing student understanding?}} \\
Academic Papers   & 72  & 70 (97.22\%) \\
Movie Plots       & 51  & 51 (100.0\%) \\
News Articles     & 60  & 58 (96.67\%) \\
Song Lyrics       & 51  & 48 (94.12\%) \\
\textbf{Overall}  & 234 & 227 (97.01\%) \\
\midrule
\multicolumn{3}{l}{\textbf{Question 2: Answerable using provided context?}} \\
Academic Papers   & 72  & 70 (97.22\%) \\
Movie Plots       & 51  & 51 (100.0\%) \\
News Articles     & 60  & 59 (98.33\%) \\
Song Lyrics       & 51  & 48 (94.12\%) \\
\textbf{Overall}  & 234 & 228 (97.44\%) \\
\midrule
\multicolumn{3}{l}{\textbf{Question 3: Avoids requiring extra knowledge?}} \\
Academic Papers   & 72  & 70 (97.22\%) \\
Movie Plots       & 51  & 51 (100.0\%) \\
News Articles     & 60  & 59 (98.33\%) \\
Song Lyrics       & 51  & 48 (94.12\%) \\
\textbf{Overall}  & 234 & 228 (97.44\%) \\
\bottomrule
\end{tabular}
\caption{Breakdown of Quiz Validation Results Across Each Domain for Three Evaluation Questions.}
\label{tab:expanded_evaluation_summary}
\end{table}

\section{Prompt Templates and Generated Samples}
\label{app:prompt_templates_static}
Table \ref{tab:combined_prompt_legend} provides a legend for prompts used in the static and dynamic settings of our study.

\begin{table}[h!]
    \centering
    \scalebox{0.75}{
    \begin{tabular}{l|c|c}
    \toprule
    \textbf{Objective} & \textbf{Reference} & \textbf{Setting} \\
    \midrule
        Lesson Generation & Listing \ref{lst:lesson_generate} & Static \\
        Quiz Generation & Listing \ref{lst:quiz_generate} & Static \\
        Quiz Generation (Images) & Listing \ref{lst:quiz_generate_image} & Static \\
        Student w/o Lesson & Listing \ref{lst:student_wo_lesson} & Static \\
        Student w/ Lesson & Listing \ref{lst:student_w_lesson} & Static \\
        Teacher w/o Lesson & Listing \ref{lst:teacher_wo_lesson} & Static \\
        Teacher w/ Lesson & Listing \ref{lst:teacher_w_lesson} & Static \\
        Student Question & Listing \ref{lst:student_prompt} & Dynamic \\
        Teacher Answer & Listing \ref{lst:teacher_prompt} & Dynamic \\
        Student Summarize Conversation & Listing \ref{lst:student_summarize} & Dynamic \\
    \bottomrule
    \end{tabular}
    }
    \caption{Legend for prompts used in the various stages of our study, including both static and dynamic experiments.}
    \label{tab:combined_prompt_legend}
\end{table}

\begin{figure*}
\begin{minipage}{\textwidth}
\begin{lstlisting}[language=, caption=Lesson Generation Prompt given Concept. We list the different prompts used for different domains in the same listing for brevity., label=lst:lesson_generate]
User:
    Movie Plots: "Prepare the student comprehensively for any quiz on this movie plot, by explaining its storyline, character arcs, themes, and significant scenes. Your explanation should cover all essential aspects, enabling the student to confidently answer questions on any part of the movie."
    
    Images: "Equip the student for any quiz on this image by providing a detailed analysis of its elements, composition, and context. Highlight the key features and underlying messages, ensuring the student can address questions related to any aspect of the image."
    
    Academic Papers: "Enable the student to excel in any quiz on this academic paper by summarizing its objectives, methodology, findings, and significance. Your summary should comprehensively cover the paper's content, preparing the student to tackle questions on any part of the study."
    
    News Articles: "Prepare the student for any quiz on this news article by outlining the main events, key figures, and the article's context. Ensure your summary is thorough, allowing the student to respond to questions on any detail of the article."
    
    Song Lyrics: "Equip the student for any quiz on these song lyrics by dissecting the narrative, themes, and expressive techniques used. Provide a complete understanding, enabling the student to engage with questions on any aspect of the lyrics."

    {concept}
\end{lstlisting}
\end{minipage}
\end{figure*}

\begin{figure*}
\begin{minipage}{\textwidth}
\begin{lstlisting}[language=, caption=Quiz generation prompt for text domains, label=lst:quiz_generate]
You are an expert educational content creator specializing in crafting challenging multiple-choice questions that require specific contextual knowledge to answer correctly. Your task is to generate 3 multiple-choice questions each at middle-school, college, and graduate levels based on the provided context. Each question should have 4 options (A, B, C, D).

**Goals:**

- **Context Dependency:** Ensure that the correct answer can only be identified using specific information from the provided context.
- **Strong Distractors:** Create plausible distractors that are carefully crafted to appear correct to someone without the context, making the question unanswerable without it.
- **Difficulty Levels:** Questions should be designed to challenge the intended academic level, progressively increasing in complexity from middle-school to graduate level.

---

**Instructions for Each Academic Level:**

**1. Middle-School Level:**

- **Question Focus:**
  - Basic recall of key facts, definitions, or direct observations from the context.
  - Questions should be straightforward, encouraging foundational understanding.
- **Distractor Design:**
  - Distractors should reflect common misconceptions or mix-ups related to the context.
  - Use simple language appropriate for middle-school students.
  - Make distractors plausible by including slight alterations or errors in details.

**2. College Level:**

- **Question Focus:**
  - Emphasize conceptual understanding, interpretations, and logical relationships within the context.
  - Explore main ideas, causes, effects, motivations, and relationships.
- **Distractor Design:**
  - Distractors must be closely aligned with the correct answer but contain subtle inaccuracies or misinterpretations.
  - Use ambiguity in language or phrasing to make multiple options appear correct.
  - Include logical misinterpretations or overemphasis on secondary details.
  - Ensure the question is difficult to answer correctly without careful analysis of the context.

**3. Graduate Level:**

- **Question Focus:**
  - In-depth analysis, symbolic or thematic interpretation, and synthesis of ideas from the context.
  - Encourage analysis of underlying theories, abstract connections, or advanced problem-solving.
- **Distractor Design:**
  - Distractors must involve subtle conceptual misalignments or alternative interpretations.
  - Introduce complex, layered reasoning or speculation grounded in the context.
  - Avoid simple factual errors; frame distractors as plausible alternatives requiring expert-level understanding to evaluate.
  - Ensure all options are so plausible that the question is unanswerable without deep knowledge of the context.

---

**Formatting Guidelines:**

- **Number the questions from 1 to 9.**
- **For each question, strictly follow this format:**

Question [number]: [Question text]
A) Option A
B) Option B
C) Option C
D) Option D
Correct Answer: [A/B/C/D]
Difficulty Level: [Middle-School/College/Graduate]

- **Do not include any explanations or additional information beyond what is specified.**

---

**Here is the context:**

{plot}

Please generate the questions as per the instructions above.
\end{lstlisting}
\end{minipage}
\end{figure*}

\begin{figure*}
\begin{minipage}{\textwidth}
\begin{lstlisting}[language=, caption=Quiz generation prompt for Images domain, label=lst:quiz_generate_image]
Generate 5 multiple-choice questions based on the provided image, each with 4 options (A, B, C, D). After each question, immediately provide the correct answer, preceded by 'Correct Answer: '. The format should be strictly followed for each question and answer pair. Here is an example of how each question and answer should be formatted:

Question 1: [Question text]
A) Option A
B) Option B
C) Option C
D) Option D
Correct Answer: A

Please adhere to this format for all 5 questions and their corresponding answers.
\end{lstlisting}
\end{minipage}
\end{figure*}

\begin{figure*}
\begin{minipage}{\textwidth}
\begin{lstlisting}[language=, caption=Prompt for student w/o lesson static evaluation, label=lst:student_wo_lesson]
The following is a quiz about "{{ concept_name }}". Choose the best option that answers the question.


- Do not output anything other than the option as A/B/C/D
- Your answer has to be one of the options. If unsure, pick your best guess.

Question: {{ question }}
Options:
A. {{ option_A }}
B. {{ option_B }}
C. {{ option_C }}
D. {{ option_D }}
\end{lstlisting}
\end{minipage}
\end{figure*}

\begin{figure*}
\begin{minipage}{\textwidth}
\begin{lstlisting}[language=, caption=Prompt for student w/ lesson static evaluation, label=lst:student_w_lesson]
Given the following lesson from a teacher about "{{ concept_name }}", choose the best option that answers the question.

- Do not output anything other than the option as A/B/C/D
- Your answer has to be one of the options. If unsure, pick your best guess.

Lesson:
{{ lesson }}

Question: {{ question }}
Options:
A. {{ option_A }}
B. {{ option_B }}
C. {{ option_C }}
D. {{ option_D }}
\end{lstlisting}
\end{minipage}
\end{figure*}

\begin{figure*}
\begin{minipage}{\textwidth}
\begin{lstlisting}[language=, caption=Prompt for teacher w/o lesson static evaluation, label=lst:teacher_wo_lesson]
Given the following context and lesson from a teacher about "{{ concept_name }}", choose the best option that answers the question.

- Do not output anything other than the option as A/B/C/D
- Your answer has to be one of the options. If unsure, pick your best guess.

Context:
{{ context }}

Question: {{ question }}
Options:
A. {{ option_A }}
B. {{ option_B }}
C. {{ option_C }}
D. {{ option_D }}
\end{lstlisting}
\end{minipage}
\end{figure*}

\begin{figure*}
\begin{minipage}{\textwidth}
\begin{lstlisting}[language=, caption=Prompt for teacher w/ lesson static evaluation, label=lst:teacher_w_lesson]
Given the following context about "{{ concept_name }}", choose the best option that answers the question.

- Do not output anything other than the option as A/B/C/D
- Your answer has to be one of the options. If unsure, pick your best guess.

Context:
{{ context }}

Lesson:
{{ lesson }}

Question: {{ question }}
Options:
A. {{ option_A }}
B. {{ option_B }}
C. {{ option_C }}
D. {{ option_D }}
\end{lstlisting}
\end{minipage}
\end{figure*}

\begin{figure*}
\begin{minipage}{\textwidth}
\begin{lstlisting}[language=, caption=Jinja-style prompt for student in dynamic conversation, label=lst:student_prompt]
To learn more about the concept "{{ concept }}" from the "{{ domain }}" domain, you can ask a teacher questions about its key aspects, characteristics, and underlying principles.

You are given the following information:
- **Concept Name**: "{{ concept }}"
{% if lesson -%}
- **Lesson (Current Understanding)**: 
"{{ lesson }}"
{%- endif %}
{% if quiz_performance -%}
- **Quiz Performance**: {{ quiz_performance }}
{%- endif %}
{% if previous_questions -%}
- **Previous Questions Asked**: You have already asked the following questions:
{% for question in previous_questions -%}
  - {{ question }}
{%- endfor %}
{%- endif %}

Additional Instructions:
- **You have a total of {{ total_questions }} questions to ask about "{{ concept }}"** of which you have asked {{ asked_questions }} questions so far.
- **Ask only one question at a time and wait for the teacher's response before asking the next question.**
- Do not list all your questions at once.
- Ask diverse questions covering its origin, purpose, structure, function, context, and significance.
- Include specific questions about key details such as themes, motifs, emotions, processes, relationships, and examples relevant to the concept.
- Ensure questions are varied, thorough, and cover all facets of the concept.
- Ask **only** one question at a time to maintain focus and clarity.
- Think creatively if you run out of questions, aiming for in-depth, insightful inquiries!
- Your response should not contain any other information besides the **one** question.

Question:
\end{lstlisting}
\end{minipage}
\end{figure*}

\begin{figure*}
\begin{minipage}{\textwidth}
\begin{lstlisting}[language=, caption=Prompt for teacher answer in dynamic conversation, label=lst:teacher_prompt]
You are a teacher tasked with answering a student's question. You have access to a reliable source, `concept_knowledge`, which contains relevant information. Your goal is to provide a clear, concise answer to the student's `question` using **only** the information from `concept_knowledge`.

### Process:

1. **Understand the Question**:
   - Identify what the student is asking.

2. **Find Relevant Information**:
   - Search `concept_knowledge` for the necessary details.
   - Address multiple parts of the question if applicable.
   - Do not copy verbatim; summarize the key points.

3. **Formulate the Response**:
   - Answer the question directly and clearly, based only on `concept_knowledge`.
   - Keep the response concise and focused on the student's needs.
   - Do not venture into information beyond the scope of `concept_knowledge` even if the student probes.

4. **Check the Response**:
   - Ensure the answer addresses the question and stays within the bounds of `concept_knowledge`.
   - If necessary, revise to make the response clearer and more concise.

### Final Response:
- Provide a direct answer to the question using information present in or closely relevant to `concept_knowledge` **only**.
- Do not reference `concept_knowledge` in your answer.
- Ensure the response is complete and accurate.
- Keep your responses concise and focused.

### Input:
- **concept_knowledge**: "{{ concept_knowledge }}"
- **question**: "{{ question }}"
\end{lstlisting}
\end{minipage}
\end{figure*}

\begin{figure*}
\begin{minipage}{\textwidth}
\begin{lstlisting}[language=, caption=Prompt for student conversation summarization in dynamic conversation, label=lst:student_summarize]
{%- if lesson is not none -%}
Lesson: 
{{lesson}}
{%- endif %}

{% for content in conversation %}
{{ content['role'].title() }}: {{ content['content'] }}
{%- endfor -%}
\end{lstlisting}
\end{minipage}
\end{figure*}

\end{document}